\documentclass[sigconf]{acmart}

\AtBeginDocument{%
  \providecommand\BibTeX{{%
    \normalfont B\kern-0.5em{\scshape i\kern-0.25em b}\kern-0.8em\TeX}}}

\copyrightyear{2023}
\acmYear{2023}
\setcopyright{rightsretained}
\acmConference[HRI '23]{Proceedings of the 2023 ACM/IEEE International Conference on Human-Robot Interaction}{March 13--16, 2023}{Stockholm, Sweden}
\acmBooktitle{Proceedings of the 2023 ACM/IEEE International Conference on Human-Robot Interaction (HRI '23), March 13--16, 2023, Stockholm, Sweden}
\acmDOI{10.1145/3568162.3576989}
\acmISBN{978-1-4503-9964-7/23/03}

\usepackage{booktabs} 
\usepackage{tikz} 
\usepackage{makecell}
\usepackage{amsfonts}
\usepackage{amsmath,stmaryrd,mathtools}

\usepackage{caption}
\usepackage{subcaption}
\usepackage{hyperref}       
\usepackage{url}
\usepackage{wrapfig}

\newcommand{\reward}{R}
\newcommand{\weight}{\ensuremath{\theta}}

\newcommand{\traj}{\ensuremath{\xi}}

\usepackage{color}

\begin{document}

\title{SIRL: Similarity-based Implicit Representation Learning}


\author{Andreea Bobu}
\authornote{Both authors contributed equally to this research. 
This research is supported by the Office of Naval Research (ONR) Young Investigator Award, the NSF National Robotics Initiative, NSF Career, the Weill Neurohub, and the Apple AI/ML Fellowship.
We thank Sid Reddy and Andrea Bajcsy for their feedback.}
\email{abobu@berkeley.edu}
\affiliation{%
  \institution{University of California, Berkeley}
  \country{United States of America}
}

\author{Yi Liu}
\authornotemark[1]
\email{yiliu77@berkeley.edu}
\affiliation{%
  \institution{University of California, Berkeley}
  \country{United States of America}
}

\author{Rohin Shah}
\email{rohinmshah@deepmind.com}
\affiliation{%
  \institution{DeepMind Research}
  \country{United Kingdom}
}

\author{Daniel S. Brown}
\email{dsbrown@berkeley.edu}
\affiliation{%
  \institution{University of California, Berkeley}
  \country{United States of America}
}

\author{Anca D. Dragan}
\email{anca@berkeley.edu}
\affiliation{%
  \institution{University of California, Berkeley}
  \country{United States of America}
}

\renewcommand{\shortauthors}{TBD}

\begin{abstract}
 
    When robots learn reward functions using high capacity models that take raw state directly as input, they need to both learn a 
    \emph{representation} for what matters in the task --- the task ``features" --- as well as how to combine these features into a single objective. 
    If they try to do both at once from input designed to teach the full reward function, it is easy to end up with a representation that contains spurious correlations in the data, which fails to generalize to new settings.
    Instead, our ultimate goal is to enable robots to identify and isolate the causal features that people actually care about and use when they represent states and behavior. 
    Our idea is that we can tune into this representation by asking users what behaviors they consider \emph{similar}: behaviors will be similar if the features that matter are similar, even if low-level behavior is different; conversely, behaviors will be different if even one of the features that matter differs. This, in turn, is what enables the robot to disambiguate between what needs to go into the representation versus what is spurious, as well as what aspects of behavior can be compressed together versus not. 
    The notion of learning representations based on similarity has a nice parallel in contrastive learning, a self-supervised representation learning technique that maps visually similar data points to similar embeddings, where similarity is defined by a designer through data augmentation heuristics. By contrast, in order to learn the representations that people use, so we can learn \emph{their} preferences and objectives, we use \emph{their} definition of similarity.
    In simulation as well as in a user study, we show that learning through such similarity queries leads to representations that, while far from perfect, are indeed more generalizable than self-supervised and task-input alternatives.
\end{abstract}


\begin{teaserfigure}
\centering
  \includegraphics[width=0.93\textwidth]{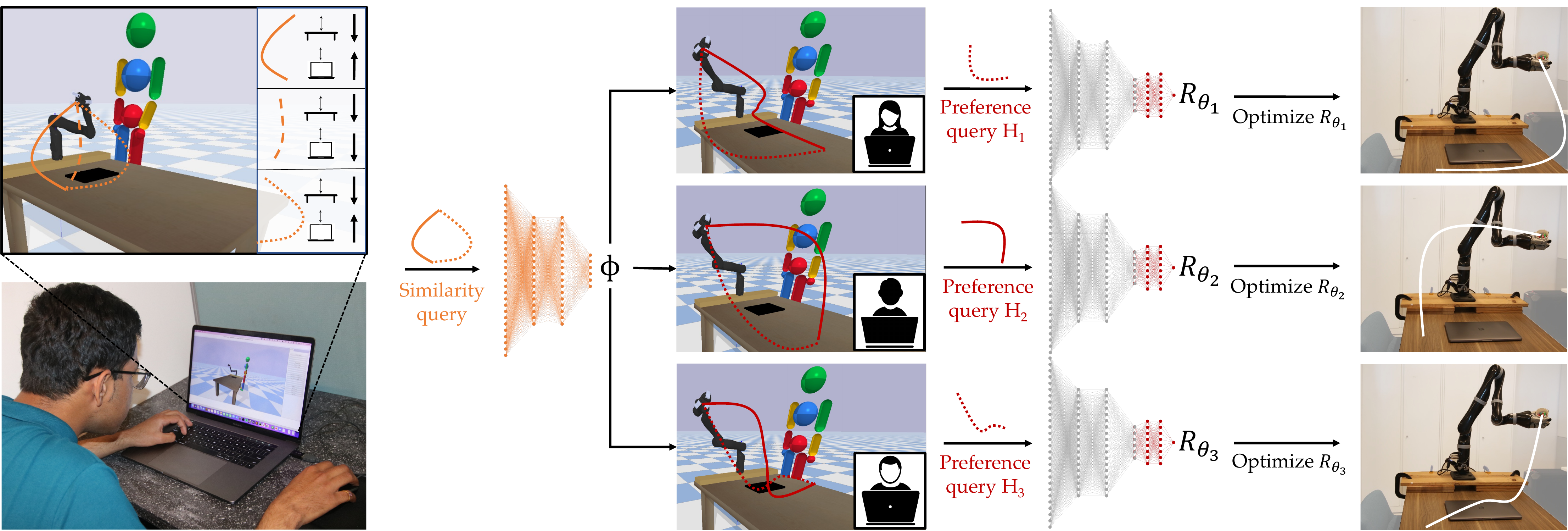}
  \caption{Our goal is to learn representations for robot behavior that capture what is salient to people, and, thus, support generalizable preference learning with low sample complexity. We propose to extract this representation by asking people trajectory similarity queries (left), where they judge which two out of three trajectories are most similar to each other. We then use the representation to learn reward functions corresponding to different people's preferences on different tasks (right).}
  \label{fig:front_fig}
\end{teaserfigure}

\maketitle

\section{Introduction}
\label{sec:intro}


Imagine waking up in the morning and your home robot assistant wants to place a steaming mug of fresh coffee on the table exactly where it knows you will sit. Depending on the context, you will have a different preference for how the robot should be doing its task. Some days it carries your favorite mug close to the table to prevent it from breaking in the case of a slip (so that it will remain your favorite mug); 
other days the steam from your delicious meal is difficult to handle for the robot's perception, so you'd want it to keep a large clearance from the table to avoid collisions. Similarly, some days you want the robot to keep your mug away from your laptop to avoid spilling on it; other days the mug only has an espresso shot so you want the robot keep it close to the laptop to prevent clutter and leave the rest of the table open for you. 

The reward function the robot should optimize changes --- whether due to variations in the task, having different users, or, as in the examples above, different contexts that are not always part of the robot state (e.g. holding the user's favorite mug and not just a regular mug). However, the \emph{representation} on top of which the reward is built, i.e the \emph{features} that are important (like the distance from the table, being above the laptop, etc.), are shared. If the robot learns this representation correctly, it can use it to obtain the right reward function, even if the task, user, and context changes. 
Meta-learning and multi-task learning methods \cite{xu2019metaIRL,gleave2018multi,nishi2020fine} learn the representation from user input meant to teach the full reward, like preference queries or demonstrations. By contrast, we propose that if learning generalizable representations is the goal, then we should ask the user for input that is specifically meant to teach the representation itself, rather than input meant to teach the full reward and hoping to extract a good representation along the way. 

But asking people to teach robots representations, rather than tasks, is not so easy. What \emph{are} the features that they care about? While some techniques advocate for people enabling users to teach each feature separately \citep{bobu2021ferl}, people may not always be able to explicate their representation and break it down into concepts that are individually teachable. 
In this work, our idea is that we can implicitly tune into the representations people use by asking them to do a proxy task of evaluating \emph{similarity} of behaviors. Behaviors will be similar if the features that matter are similar, even if low-level behavior is different; conversely, behaviors will be different if even one of the features that matter differs. This, in turn, should enable the robot to arrive at the features that matter ---  we want robots that can disambiguate between what needs to go into the representation versus what is spurious, as well as what aspects of behavior can be compressed together into a feature embedding versus kept separate. 
We thus introduce a novel type of human input to help the robot extract the person's representation: \textit{trajectory similarity queries}.
A trajectory similarity query is a triplet of trajectories that the person answers by picking the two more similar trajectories. In Figure ~\ref{fig:front_fig} (left), the person chooses the two trajectories that are close to the table and far from the laptop, even though visually they look dissimilar. This results in an (anchor, positive, negative) triplet that can be used for training a feature representation. We call this process Similarity-based Implicit Representation Learning (SIRL).

Our method has a parallel in self-supervised learning work, especially contrastive learning, where the goal is to learn a good visual representation by training from (anchor, positive, negative) triplets generated via data augmentation techniques~\cite{chen2020simple}. However, this notion of similarity is purely visual, driven by manually designed heuristics for data augmentation, and is not necessarily reflective of what users would consider similar. For instance, two images might be labeled as  visually different, when in fact their difference is only with respect to some low-level aspects that are not really relevant to the distribution of tasks people care about. This would result in representations that contain too many distractor features that are not present in the human's representation. Our method uses similarity too, but we defer to the user's judgement of similarity, with the goal of reconstructing the \emph{user's representation}.

Of course, our method is not the full answer to learning causally aligned representations. But our experiments suggest that it outperforms methods that are self-supervised, or that learn from input meant to teach the full tasks. In simulation, where we know the causal features, we show that SIRL learns representations better aligned with them, which in turn leads to learning multiple more generalizable reward functions downstream (Figure \ref{fig:front_fig}). We also present a user study where we crowdsource similarity queries from different people to learn a shared SIRL representation that better recovers each of their individual preferences. While the study results do show a significant effect, the effect size is much lower than in simulation. This is attributable in part to the interface difficulty of analyzing the robot trajectories, which means more work is needed to determine the best interfaces that enable users to accurately answer similarity queries. Moreover, some users reported struggling to trade off the different features, which means that similarity queries might not be entirely preference-agnostic. Nonetheless, our results underscore that there are gains by explicitly aligning robot and human representations, rather than hoping it will happen as a byproduct of learning rewards from standard queries.

\section{Related Work}
\label{sec:related}

\noindent\textbf{Learning from Human Input.}
Human-in-the-loop learning is a well-established paradigm where the robot uses human input to infer a policy or reward function capturing the desired behavior. In imitation learning, the robot learns a policy that essentially copies human demonstrations~\citep{osa2018algorithmic}, a strategy that typically doesn't generalize well outside the training regime~\citep{levine2020offline}. Meanwhile, inverse reinforcement learning (IRL) uses the demonstrations to extract a reward function capturing \textit{why} a specific behavior is desirable, thus better generalizing to unseen scenarios~\citep{abbeel2004apprenticeship}. Recent research goes beyond demonstrations, utilizing other types of human input for reward learning, such as corrections~\citep{bajcsy2017phri}, comparisons~\citep{wirth2017survey}, or rankings~\citep{brown2019extrapolating}. Unless explicitly designed for, these methods learn a latent representation implicitly from the respective human input. We seek to instead explicitly learn a preference-agnostic latent space that can be used for downstream tasks like reward learning. We focus on learning models of human reward functions via pairwise preference queries~\citep{wirth2017survey}, but we believe the latent space we learn can be useful for learning from any of the above types of feedback.

\smallskip
\noindent \textbf{Representation and Similarity Learning.}
Common representation learning approaches are unsupervised~\cite{chen2016infogan,Higgins2017betaVAELB,chen2018betaTCVAE} or self-supervised~\cite{Doersch2015UnsupervisedVR,pathak2018zero,aytar2018playing,laskin2020curl}, but because they are purposefully designed to bypass human supervision, the learned embedding does not necessarily correspond to features the person cares about. 
Prior work leverages task labels~\cite{chen2021DVD} or trajectory rankings~\cite{brown2020brex} to learn latent spaces that help identify specific goals or preferences.
By contrast, we focus on learning task-agnositic measures of feature similarity that are useful for learning multiple preferences.
Some work looks at having people interactively select features from a pre-defined set~\cite{cakmak2012questions,bullard2018featureselection, hoai2019refinement} or teach task-agnostic features sequentially via kinesthetic feature demonstrations~\cite{bobu2022inducing} or active learning techniques~\cite{kulick2013symbols, hayes2014constraints,bobu2022perceptual}.
We instead focus on fully learning a lower-dimensional feature representation all-at-once, rather than one at a time. Furthermore, rather than relying on the human to provide physical demonstrations for learning a good feature space~\cite{bobu2021ferl,bobu2022inducing}, we propose a more accessible and general form of human feedback: showing the user triplets of trajectories and simply asking them to label which two trajectories are the most similar. Triplet losses have been widely used to learn similarity models that capture how humans perceive objects~\cite{agarwal2007generalized,tamuz2011adaptively,mcfee2011learning,cagatay2014kernels,ehsan2015kernel}; however, to the best of our knowledge, we are the first to use a triplet loss to learn a general, task-agnostic similarity model of how humans perceive trajectories. 

\smallskip
\noindent \textbf{Meta- and Multi-Task Reward Function Learning.}
To learn multiple models of human reward functions, prior work has proposed clustering demonstrations and learning a different reward function for each cluster~\cite{dimitrakakis2011bayesian,babes2011apprenticeship,choi2012nonparametric}; however, these methods require a large number of demonstrations and do not adapt to new reward functions. 
Meta-learning~\cite{finn2017maml} seeks to learn a reward function initialization that enables fast fine-tuning at test time~\cite{xu2019learning,yu2019meta,huang2021meta,seyed2019smile}. Multi-task reward learning approaches pretrain a reward function on multiple human intents and then fine-tune the reward function at test time~\cite{gleave2018multi,nishi2020fine}. This has been shown to be more stable and scalable than meta-learning approaches~\cite{mandi2022effectiveness}, but still requires curating a large set of training environments. By contrast, we do not assume any knowledge of the test-time task distribution \textit{a priori} and do not require access to a population of different reward functions during training. Rather, we focus on learning a task-agnostic feature representation that can be utilized for down-stream reward learning tasks. In particular, we test our learned representation on the down-stream task of learning models of human reward functions via pairwise preference queries over trajectories~\cite{wirth2017survey,sadigh2017active,biyik2018batch,li2021roial}.

\section{Method}
\label{sec:method}

We present our method for learning preference-agnostic representations from trajectory similarity queries. Our intuition is that if a human judges two behaviors to be similar, then their representations should also be similar. Since directly asking if two trajectories are similar is difficult without an explicit threshold, we instead present the human with a triplet of trajectories and ask them to pick the two most similar (or, equivalently, the most dissimilar one). We use the human's answers to train the representation such that similar trajectories have embeddings that are close and dissimilar trajectories map to embeddings far apart. The robot then uses this latent space as a shared representation for downstream preference learning tasks with multiple people, each with different preferences.

\subsection{Preliminaries}


We define a trajectory $\xi$ as a sequence of states
, and denote the space of all possible trajectories by $\Xi$. 
The human's preference over trajectories is given by a reward function $R: \Xi \mapsto \mathbb{R}$ that is unobserved by the robot and must be learned from human interaction. 
The robot reasons over a parameterized approximation of the reward function $R_\weight$, where $\weight$ represents the parameters of a neural network. To learn $\weight$, the robot collects human preference labels over trajectories~\citep{wirth2017survey,christiano2017preferences} and seeks to find parameters $\theta$ that maximize the likelihood of the human input. The robot can then use the learned reward function to score trajectories during motion planning in order to align its behavior with a particular human's preferences. 
We focus on explicitly using human input to first learn a good representation and then use that representation for downstream reward learning, rather than using reward-specific human input (e.g., preferences or demonstrations) to implicitly learn the representation at the same time as the reward function.


\subsection{Training the Feature Representation via Trajectory Similarity Queries}
We seek to train a latent space that is useful for multiple downstream preference learning tasks. To do this, we propose learning a preference-agnostic model of human similarity. 
One way to learn such a model would be to ask users to judge whether two trajectories are similar or not; however, humans are better at giving relative rather than binary or quantitative assessments of similarity~\cite{kendall1948rank,stewart2005absolute}. 
Thus, rather than asking users to use some internal threshold or scoring mechanism to quantitatively measure similarity, we instead focus on qualitative trajectory similarity queries. 
We present the user with a visualization of three trajectories and ask them to pick the two most similar ones (equivalently the most dissimilar one). The human's queries form a data set $\mathcal{D}_{sim} = \{(\xi_{P_1}^i, \xi_{P_2}^i, \xi_N^i)\}$, where $\xi_{P_1}^i$ and $\xi_{P_2}^i$ are the trajectories that are most similar and $\xi_N^i$ is the trajectory most dissimilar to the other two.

We can interpret similarity (or dissimilarity) as a distance function, so we define the distance between two trajectories as the $L_2$ feature distance: $d(\xi_1, \xi_2) = \|\phi(\xi_1)  - \phi(\xi_2) \|_2^2$.
Given a dataset of trajectory similarity queries $\mathcal{D}_{sim}$, we use the triplet loss~\citep{balntas2016triplet}:
\begin{equation}
    \mathcal{L}_{trip}(\xi_A,\xi_P,\xi_N) = \max (d(\xi_A, \xi_P) - d(\xi_A, \xi_N) + \alpha, 0 ) \enspace,
\end{equation}
a form of contrastive learning where $\xi_A$ is the anchor, $\xi_P$ is the positive example, $\xi_N$ is the negative example, and $\alpha \geq 0$ is a margin between positive and negative pairs. However, because our queries do not contain an explicit anchor, our final loss is as follows:
\begin{equation}
    \mathcal{L}_{sim}(\phi) = \sum_{i=1}^{|\mathcal{D}_{sim}|}  \mathcal{L}_{trip}(\xi_{P_1}^i,\xi_{P_2}^i,\xi_N^i) +  \mathcal{L}_{trip}(\xi_{P_2}^i,\xi_{P_1}^i,\xi_N^i) \enspace.
    \label{eq:SIRLloss}
\end{equation}
We train a similarity embedding function $\phi: \Xi \mapsto \mathbb{R}^d$ that minimizes the above similarity loss, where $d$ is the representation dimensionality. The intuition is that optimizing this loss should push together the embeddings of similar trajectories and push apart the embeddings of dissimilar trajectories.
Before training the representation with the loss in Eq. \eqref{eq:SIRLloss}, we may also pre-train it using unsupervised learning~\citep{kingma2014vae}, which we experiment with in Sec. \ref{sec:experiments}. 

\subsection{Using SIRL for Reward Learning}

Given a learned embedding $\phi$, we can use it for learning models of specific user preferences. While we focus on learning from pairwise preferences, we note that $\phi$ can in principle be used in downstream tasks that learn from many types of human feedback~\cite{jeon2020reward}. When learning a reward function from human preferences, we show the human two trajectories, $\xi_A$ and $\xi_B$, and then ask which of these two the human prefers. We collect a data set of such preferences $\mathcal{D}_{pref} = \{(\xi_A^i, \xi_B^i, \ell^i)\}$ where $\ell^i=1$ if $\traj_A^i$ is preferred to $\traj_B^i$, denoted $\traj_A^i \succ \traj_B^i$, and $\ell^i=0$ otherwise. We interpret the human's preferences through the lens of the Bradley-Terry preference model~\cite{bradley1952rank}:
\begin{equation}
   P_\theta (\traj_A \succ \traj_B)=\frac{e^{ \reward_\weight(\phi(\traj_A))}}{e^{ \reward_\weight(\phi(\traj_A))}+e^{ \reward_\weight(\phi(\traj_B))}} \enspace.
   \label{eq:bradley}
\end{equation}
%
We learn the reward function with a simple cross-entropy loss:
\begin{equation}
    \mathcal{L}_{pref}(\weight) = - \sum_{i=0}^{|\mathcal{D}_{pref}|} \ell^i \log P_\theta(\traj_A^i \succ \traj_B^i) + (1-\ell^i) \log P_\theta(\traj_B^i \succ \traj_A^i) \enspace.
    \label{eq:preference}
\end{equation}

\subsection{Adapting to Different User Preferences}
We want robots that can adapt to changes to an individual user's preferences depending on the context as well as quickly adapt to new users' preferences. 
Rather than learn each preference independently by collecting a new set of human data and training a completely new reward function $R_{\weight}$, we study whether we can leverage the latent space learned by SIRL to perform more accurate and sample-efficient multi-preference learning. 
When learning a new user's preference model, $R_{\theta}$ the robot can use $\phi$ to more quickly learn the reward function $R_\theta(\phi(\xi))$. Our main idea is that because this shared latent representation $\phi$ is trained via preference-agnostic similarity queries, it is more transferable than using a multi-task or meta-learning approach, where the pre-trained network is trained using multiple, specific task objectives. Furthermore, because SIRL uses human input to train $\phi$, we hypothesize that the learned feature space will be better suited for learning human reward functions than a latent space learned via unsupervised training.

\section{Experiments in Simulation}
\label{sec:experiments}

We first investigate the quality of SIRL-trained representations and their benefits for preference learning using simulated human input in two environments with ground truth rewards and features.

\subsection{Environments}


\begin{figure}
\begin{subfigure}[b]{0.25\textwidth}
\centering
\includegraphics[width=\textwidth]{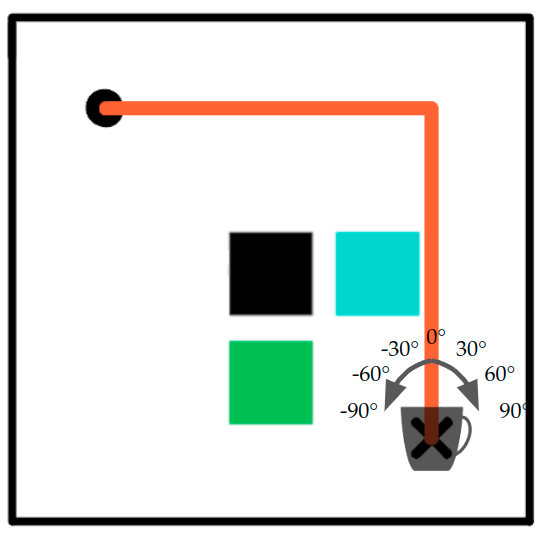}\\
\caption{GridRobot environment.}
 \label{fig:gridrobot_env}
\end{subfigure}
\begin{subfigure}[b]{0.45\textwidth}
\centering
\includegraphics[width=\textwidth]{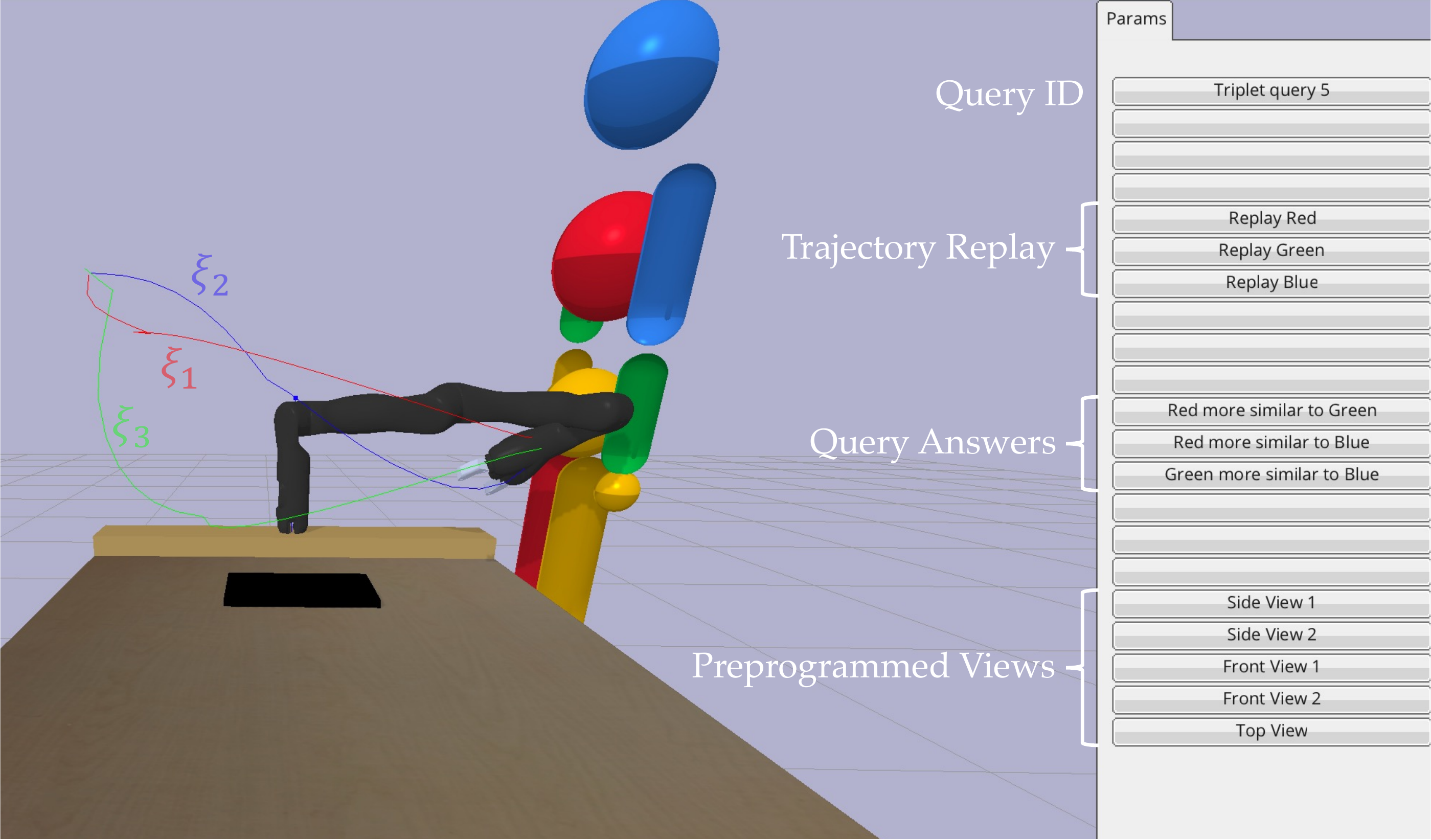}\\
\caption{JacoRobot environment and user study interface.}
 \label{fig:jacorobot_env}
\end{subfigure}
\vspace{-3mm}
\caption{Visualization of the experimental environments.}
\label{fig:environments}
\vspace{-5mm}
\end{figure}

\textbf{\textit{GridRobot}} (Figure  \ref{fig:gridrobot_env}) is a $5$-by-$5$ gridworld with two obstacles and a laptop (the blue, green, and black boxes).
Trajectories are sequences of 9 states with the start and end in opposite corners. The 19-dimensional input consists of the $x$ and $y$ coordinates of each state and a discretized angle in $\{-90^{\circ}, -60^{\circ}, -30^{\circ}, 0^{\circ}, 30^{\circ}, 60^{\circ}, 90^{\circ}\}$ at the end state. The simulated human answers queries based on 4 features $\phi^*$ in this world: Euclidean distances to each object, and the absolute value of the angle orientation.

\textbf{\textit{JacoRobot}} (Figure  \ref{fig:jacorobot_env}) is a pybullet~\citep{coumans2019} simulated environment with a 7-DoF Jaco robot arm on a tabletop, with a human and laptop in the environment. Trajectories are length 21, and each state consists of 97 dimensions: the $xyz$ positions of all robot joints and objects, and their rotation matrices. This results in a 2037-dimensional input space, much larger than for GridRobot. The 4 features of interest $\phi^*$ for the simulated human are: a) \textit{table} --- distance of the robot's End-Effector (EE) to the table;
b) \textit{upright} --- EE orientation relative to upright, to consider whether objects are carried upright;
c) \textit{laptop} --- $xy$-plane distance of the EE to a laptop, to consider whether the EE passes over the laptop at any height;
d) \textit{proxemics}~\cite{mumm2011human} --- proxemic $xy$-plane distance of the EE to the human, where the EE is considered closer to the human when moving in front of the human that to their side.

In GridRobot the state space is discretized, so the trajectory space $\Xi$ can be enumerated; however, the JacoRobot state space is continuous, so we construct $\Xi$ by smoothly perturbing the shortest path trajectories from 10,000 randomly sampled start-goal pairs (see App. \ref{app:trajgen}). We generate similarity and preference queries by randomly sampling from $\Xi$.
The simulated human answers similarity queries by computing the 4 feature values for each of the three trajectories and choosing the two that were closest in the feature space. For preference queries, the simulated human computes the ground truth reward and samples the trajectory with the higher reward. The space of true reward functions (used to simulate preference labels) is defined as linear combinations of the 4 features described above. The robot is not given access to the ground-truth features nor the ground-truth reward function but must learn them from similarity and preference labels over raw trajectory observations.

\subsection{Qualitative Examples}

\begin{figure}
\centering
\includegraphics[width=0.45\textwidth]{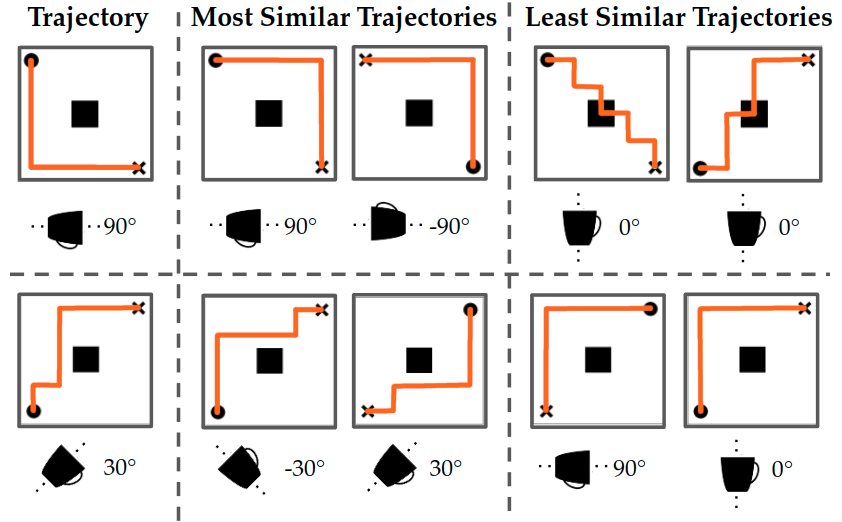}\\
\caption{SIRL picks the two most and least similar trajectories to a query trajectory. Top: trajectories are similar in features despite being dissimilar in states. Bottom: trajectories are dissimilar in features despite being close in states.}
 \label{fig:SIRL_most_similar}
 \vspace{-4mm}
\end{figure}

In Figure \ref{fig:SIRL_most_similar} we show similar and dissimilar trajectories learned by SIRL in a simplified GridRobot environment with only the laptop and the joint angle. \textit{Top}: the given trajectory stays far from the laptop and holds the cup on its side; SIRL learns that trajectories that share those features are similar, despite being dissimilar in the state-space. \textit{Bottom}: the trajectory stays close to the laptop and holds the cup at an angle; SIRL learns that trajectories that hold the cup upright and stay far from the laptop are dissimilar, despite being similar in the state-space (going up and then right).

\subsection{Experimental Setup}

\smallskip
\noindent\textbf{Manipulated Variables.}
We test the importance of user input that is designed to teach the representation by comparing SIRL with multi-task learning techniques from generic preference queries, and unsupervised representation learning. We have 4 baselines:
a) \textbf{\textit{VAE}}, which learns a representation with a variational reconstruction loss~\cite{kingma2014vae};
b) \textbf{\textit{MultiPref}}, a multi-task baseline~\cite{gleave2018multi}, where we learn the representation $\phi$ implicitly by training multiple reward functions (each with shared initial layers) via preference learning;
c) \textbf{\textit{SinglePref}}, a hypothetical method that learns from an ideal user who weighs all features equally; 
d) \textbf{\textit{Random}}, a randomly initialized embedding, which does not benefit from human data but is also immune from any spurious correlations that might be learned from biased data.
For MultiPref, we trained versions with 10 and 50 simulated human preference rewards for good coverage of the reward space.  All embeddings have the same network size: for GridRobot we used MLPs with 2 layers, 128 units each, mapping to 6 output neurons, while for JacoRobot we used 1024 units to handle the larger input space (see App. \ref{app:details}). 
For a fair comparison, we gave SIRL, SinglePref, and MultiPref equal amounts of human data for pre-training: $N$ similarity queries for SIRL, and $N$ preference queries (used for a single human for SinglePref or equally distributed amongst humans for MultiPref). We also performed ablations with and without VAE pre-training and found that SinglePref and MultiPref are better without VAE (see App. \ref{app:ablations}). 

\smallskip
\noindent\textbf{Dependent Measures.}
To test the quality of the learned representations, we use two metrics: \textit{Feature Prediction Error (FPE)} and \textit{Test Preference Accuracy (TPA)}. The \textit{\textbf{FPE}} metric is inspired by prior work that argues that good representations are linearly separable~\citep{Coates2012LearningFR,lai2019contrastive,reed2022separability}. Our goal is to measure whether the embeddings contain the necessary information to recover the 4 ground-truth features in each environment. 
We generate data sets of sampled trajectories labeled with their ground truth (normalized) feature vector $\mathcal{D}_{FPE} = \{\xi, \phi^*\}$. We freeze each embedding and add a linear regression layer on top to predict the feature vector for a given trajectory. We split $\mathcal{D}_{FPE}$ into 80\% training and 20\% test pairs, and \textit{FPE} is the mean squared error (MSE) on the test set between the predicted feature vector and the ground truth feature vector. For the human query methods, we report \textit{FPE} with increasing number of representation training queries $N$.

For \textit{\textbf{TPA}}, we test whether good representations necessarily lead to good learning of general preferences. We use the trained embeddings as the base for 20 randomly selected test preference rewards. 
For each $R_{\theta_i}$, we generate a set of labeled preference queries $\mathcal{D}_{pref}^{\theta_i} = \{\xi_A, \xi_B, l\}$, which we split into 80\% for training and 20\% for test. We train each reward model with $M$ preference queries per test reward, and we vary $M$. All preference networks have the same architecture: we take the embedding $\phi$ pre-trained with the respective method, and add new fully connected layers to learn a reward function from trajectory preference labels. For GridRobot we used MLPs with 2 layers of 128 units, and for JacoRobot we used 1024 units. We found that all methods apart from SIRL worked better with unfrozen embeddings (App. \ref{app:ablations}). 
We report TPA as the preference accuracy for the learned reward models on the test preference set, averaged across the test human preferences.

\smallskip
\noindent\textbf{Hypotheses.} We test two hypotheses: 

\textbf{H1.} Using similarity queries specifically designed to teach the representation (SIRL) leads to representations more predictive of the true features (lower \textit{FPE}) than unsupervised (VAE), implicit (MultiPref, SinglePref), or random representations. 

\textbf{H2.}
The SIRL representations result in more generalizable reward learning (higher \textit{TPA}) than unsupervised (VAE), implicit (MultiPref, SinglePref), or random representations. 

\begin{figure*}[t!]
\begin{subfigure}[b]{0.49\textwidth}
\centering
\includegraphics[width=\textwidth]{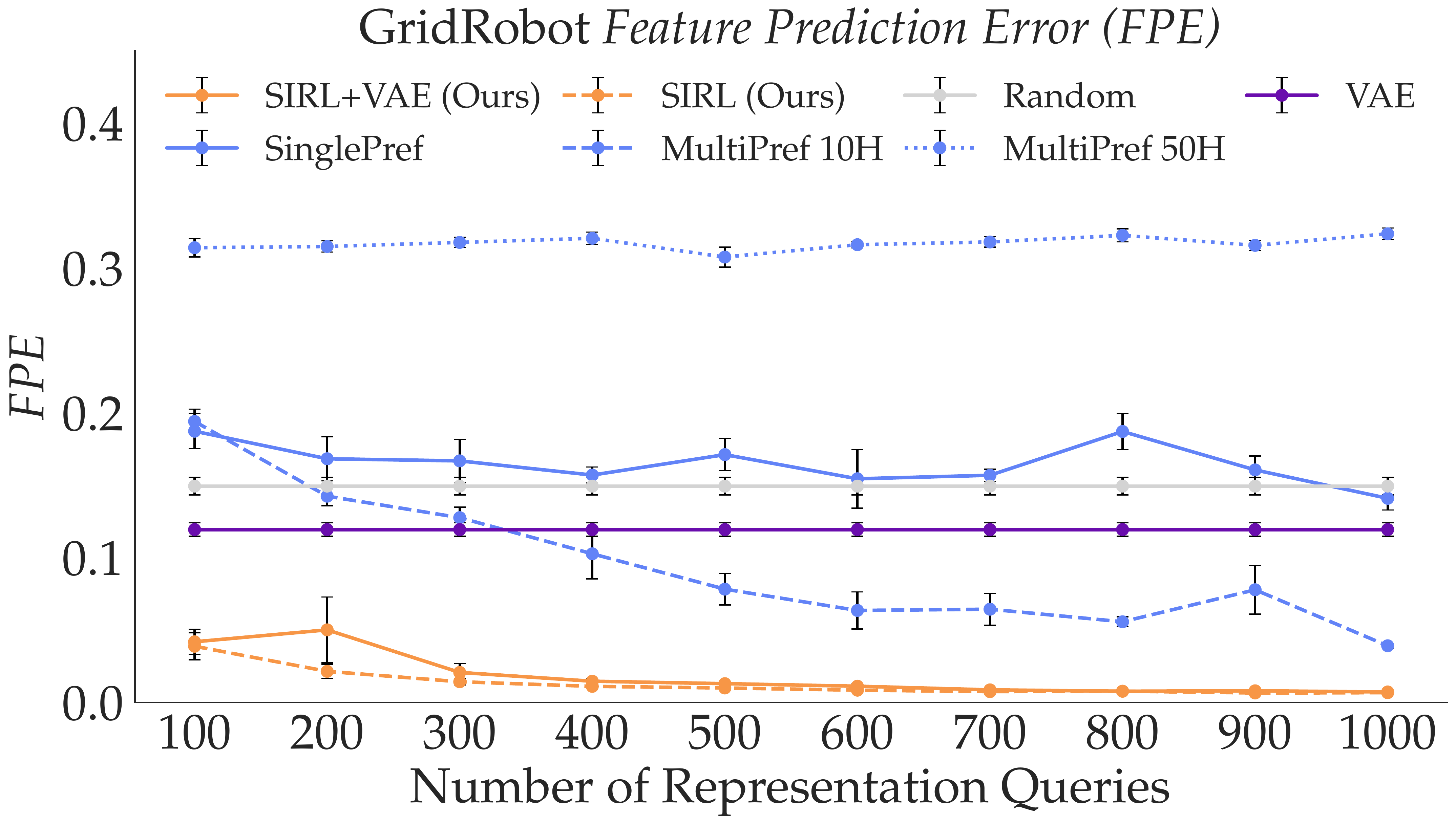}\\
 \label{fig:gridrobot_PFE}
\end{subfigure}
\begin{subfigure}[b]{0.49\textwidth}
\centering
\includegraphics[width=\textwidth]{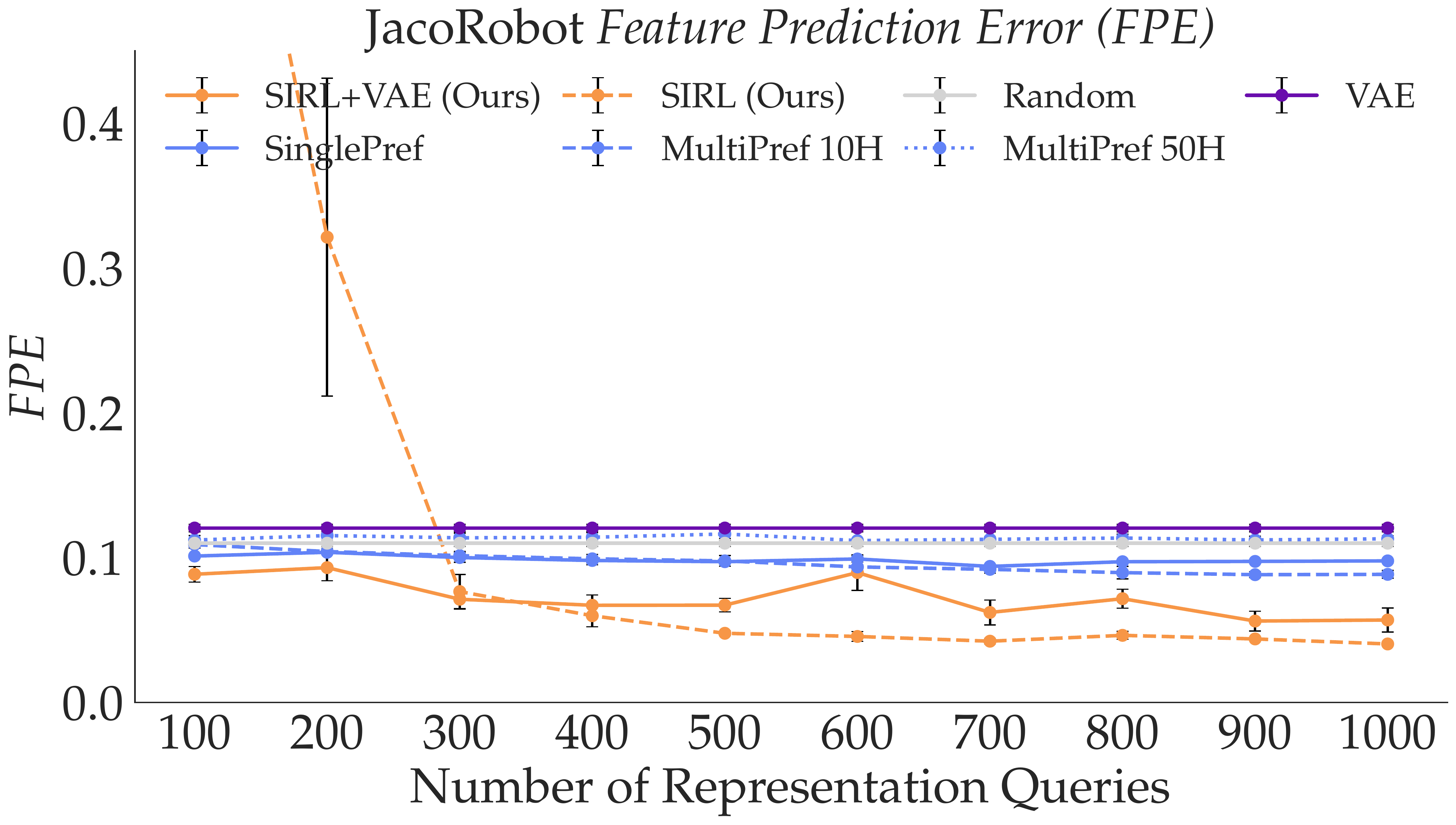}\\
 \label{fig:jacorobot_PFE}
\end{subfigure}
\caption{\textit{FPE} for the GridRobot (left) and JacoRobot (right) environments with simulated human data. With enough data, SIRL learns representations more predictive of the true features $\phi^*$ in both simple and complex environments.}
\label{fig:FPE_simulation}
\vspace{-2mm}
\end{figure*}

\subsection{Results}
In Figure \ref{fig:FPE_simulation} we show the \textit{FPE} score for both environments with varying representation queries $N$ from 100 to 1000. For GridRobot, both versions of SIRL (with or without VAE pre-training) perform similarly and outperform all baselines, supporting H1. When pre-training with preference queries, MultiPref with 10 humans performs better than SinglePref or MultiPref with 50 humans: SinglePref may be overfitting to the one human preference it has seen, while when MultiPref has to split its data budget among 50 humans it ends up learning a worse representation than Random.
There is a balance to be struck between the diversity in human training rewards covered and the amount of pre-training data each reward gets, a trade-off which SIRL avoids because similarity queries are agnostic to the particular human reward.
For the more complex JacoRobot, both versions of SIRL outperform all baselines, in line with H1, although SIRL without VAE scores better than with it.

\begin{figure*}
\begin{subfigure}[b]{0.495\textwidth}
\centering
\includegraphics[width=\textwidth]	{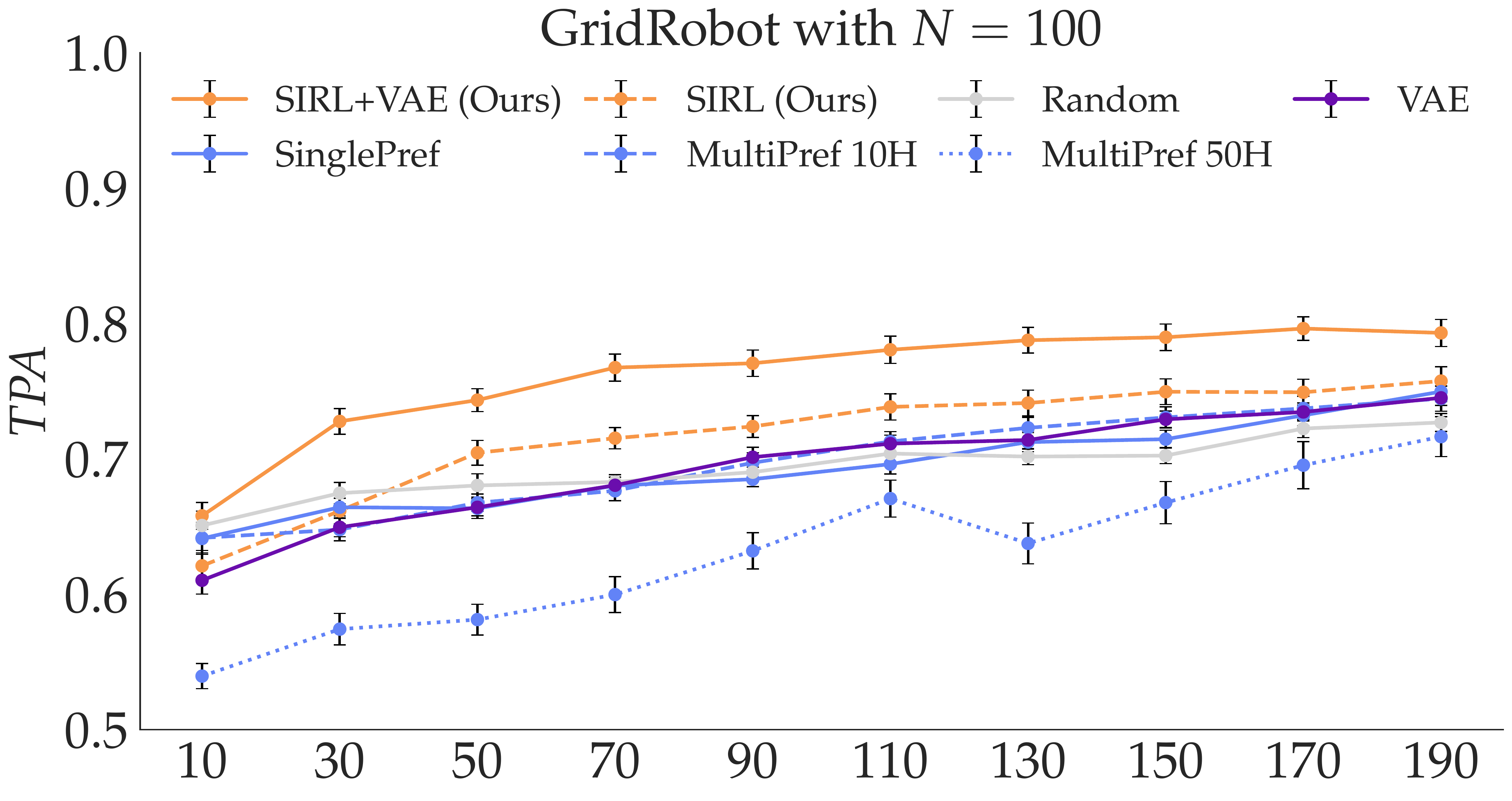}\\
 \label{fig:gridrobot_TPA_100}
\end{subfigure}
\begin{subfigure}[b]{0.495\textwidth}
\centering
\includegraphics[width=\textwidth]{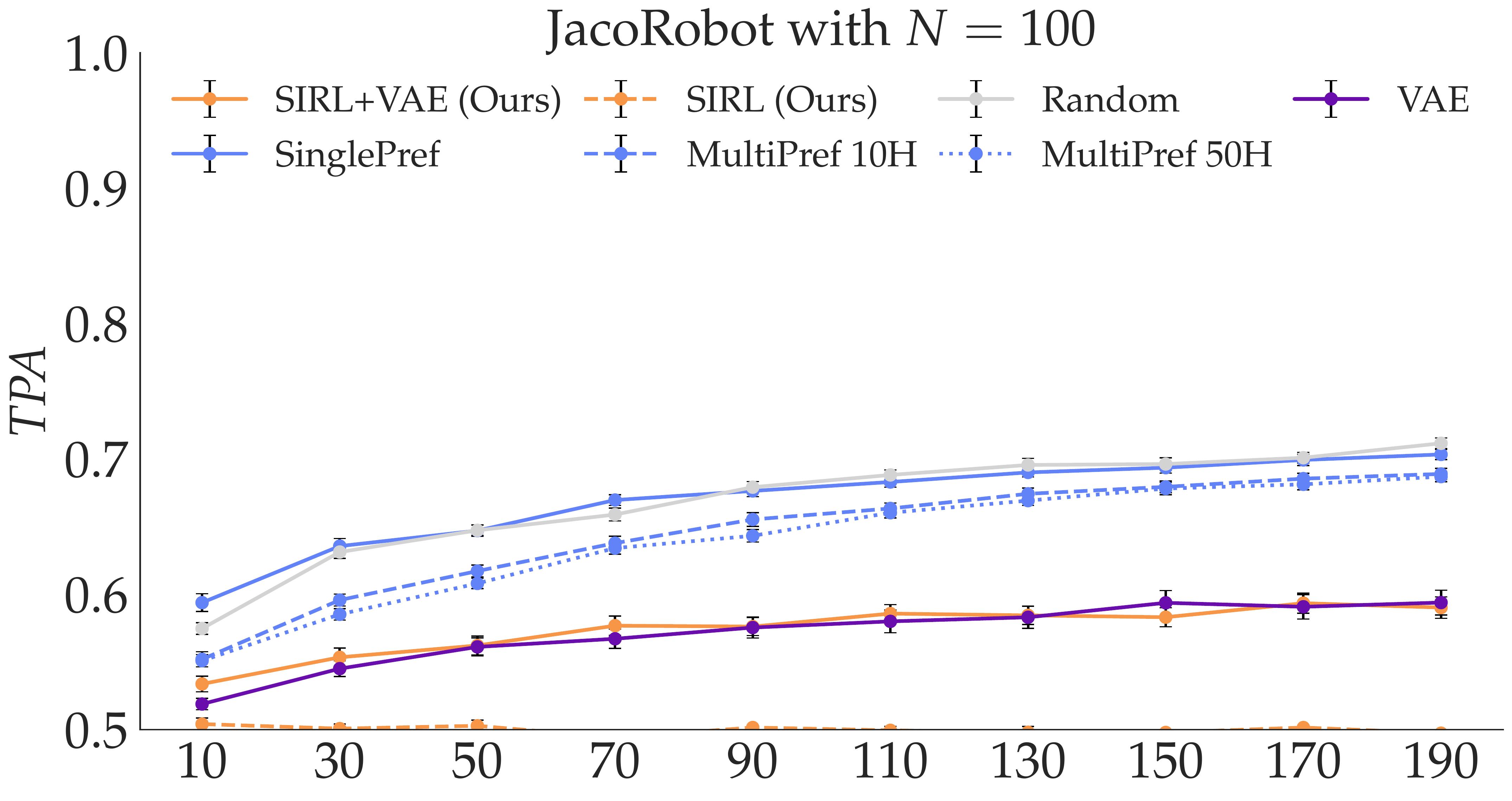}\\
 \label{fig:gridrobot_TPA_500}
\end{subfigure}
\begin{subfigure}[b]{0.495\textwidth}
\centering
\includegraphics[width=\textwidth]	{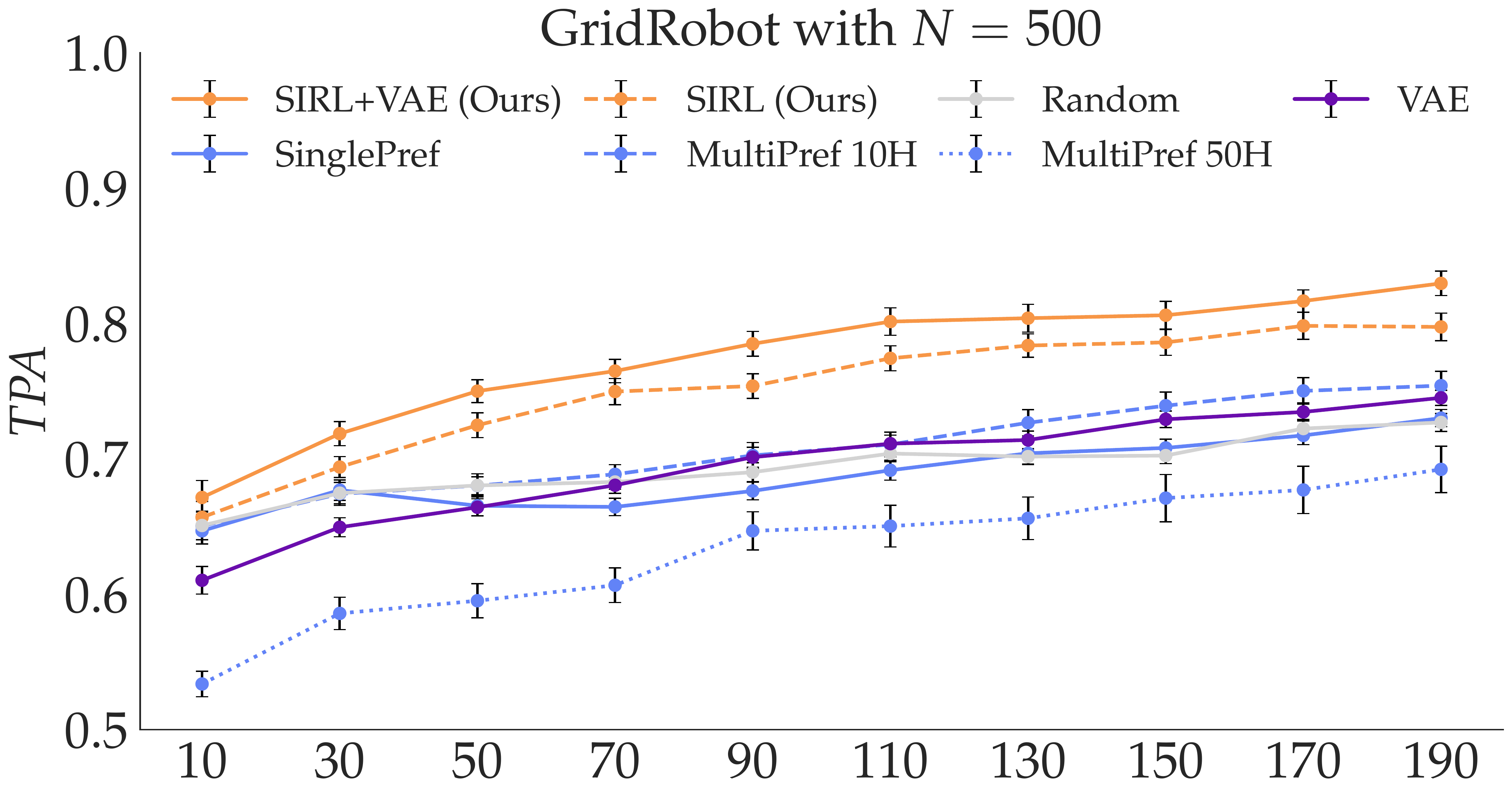}\\
 \label{fig:gridrobot_TPA_1000}
\end{subfigure}
\begin{subfigure}[b]{0.495\textwidth}
\centering
\includegraphics[width=\textwidth]	{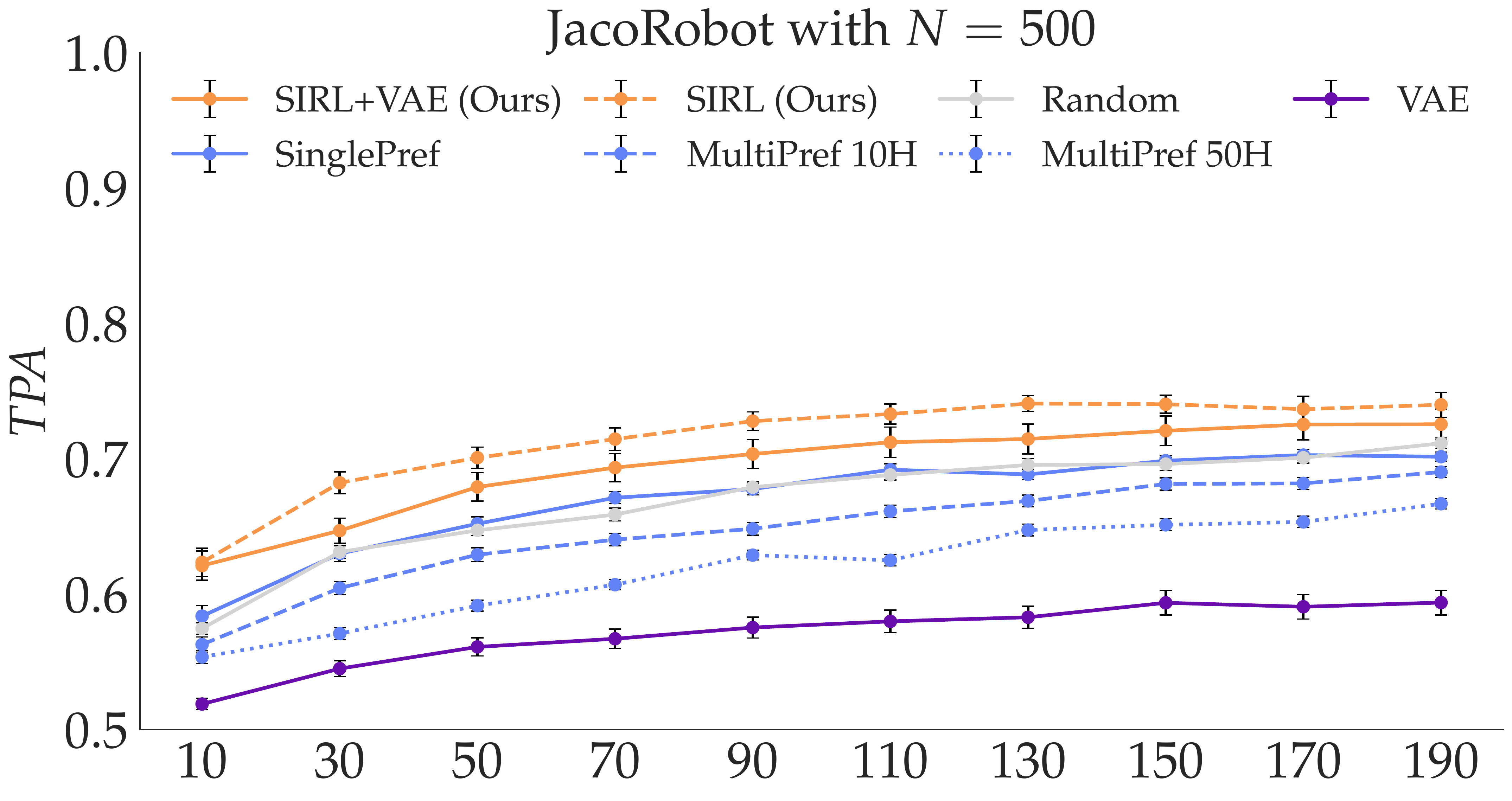}\\
 \label{fig:jacorobot_TPA_100}
\end{subfigure}
\begin{subfigure}[b]{0.495\textwidth}
\centering
\includegraphics[width=\textwidth]	{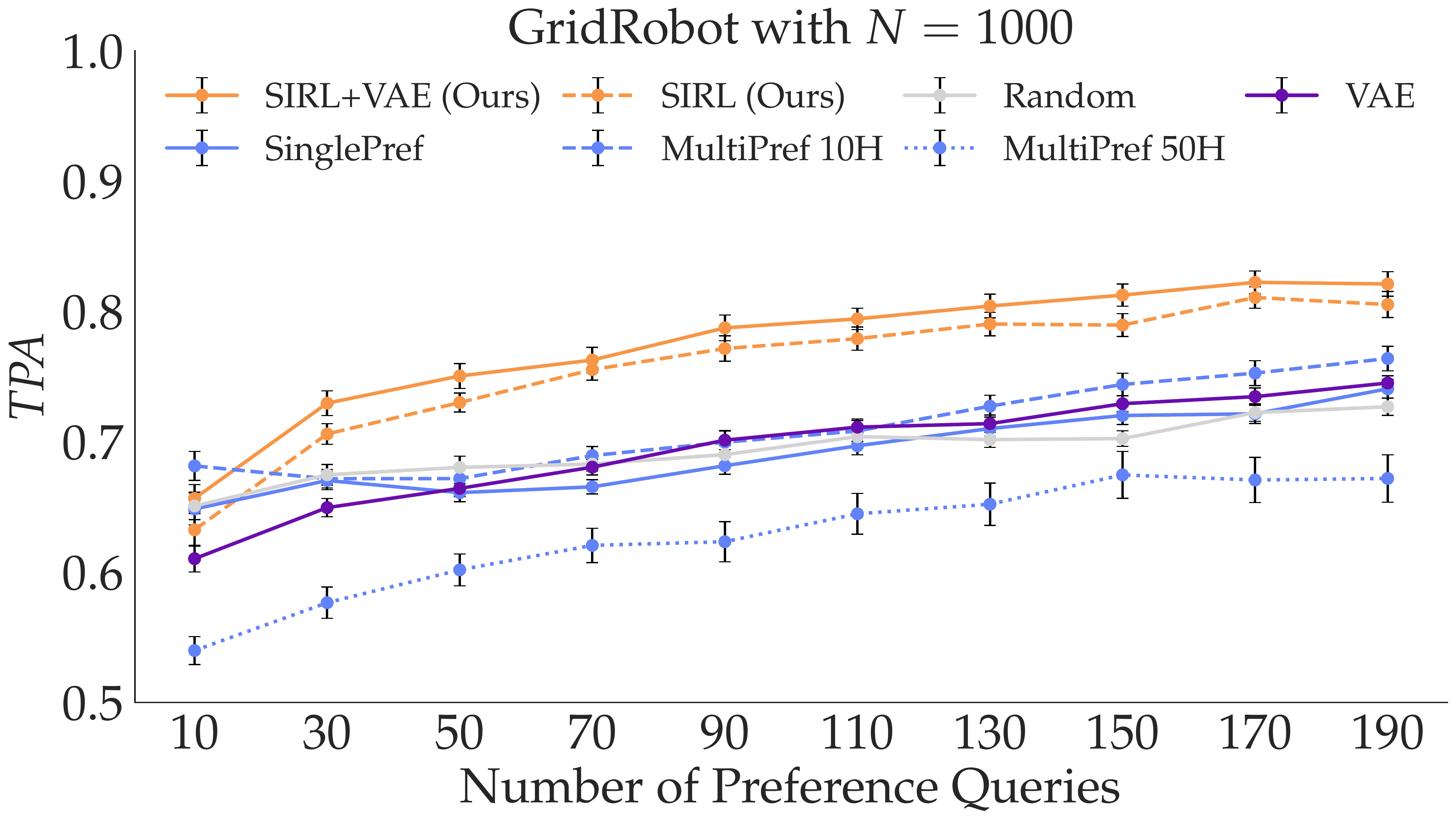}\\
 \label{fig:jacorobot_TPA_500}
\end{subfigure}
\begin{subfigure}[b]{0.495\textwidth}
\centering
\includegraphics[width=\textwidth]	{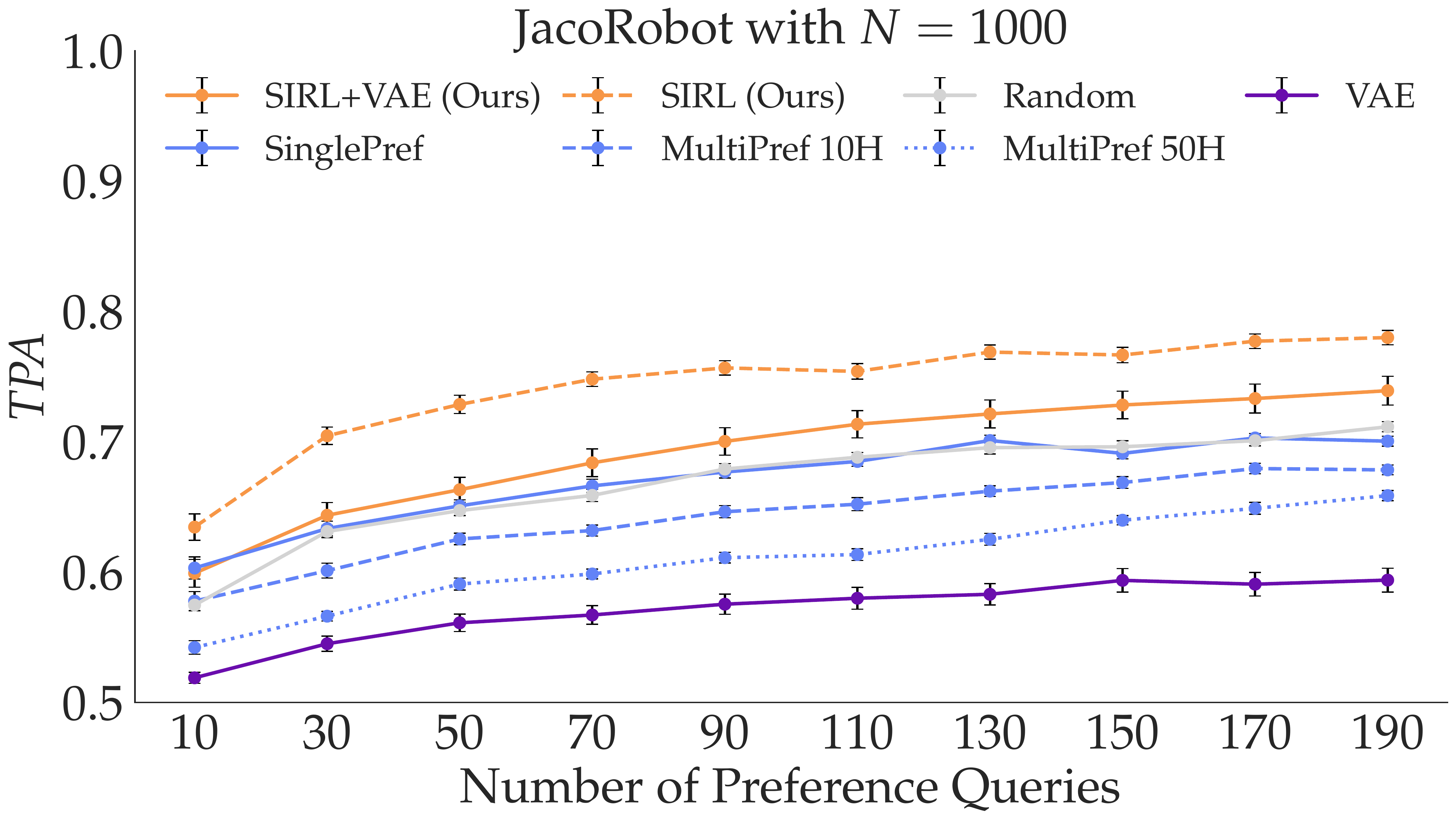}\\
 \label{fig:jacorobot_TPA_1000}
\end{subfigure}
\caption{\textit{TPA} for GridRobot (top) and JacoRobot (bottom) with simulated human data. With enough data, SIRL recovers more generalizable rewards than unsupervised, preference-trained, or random representations.}
\vspace{-2mm}
\label{fig:TPA_simulation}
\end{figure*}

In Figure  \ref{fig:TPA_simulation} we present the \textit{TPA} score for both environments with a varying amount of test preference queries $M$ from 10 to 190, and $N=100$, $500$, and $1000$. For GridRobot, each respective method performs comparably with different $N$s, suggesting that this is a simple enough environment that low amounts of representation data are sufficient. For JacoRobot, this is not the case: with just 100 queries, SIRL with VAE pre-training performs like VAE, SIRL without pre-training has random performance (since it's frozen), and the preference baselines all perform close to Random, as if they weren't trained with queries at all. For larger $N$, both versions of SIRL start performing better than the baselines, suggesting that with enough data a good representation can be learned.

Focusing on $N=1000$, our results support H2: both SIRLs outperform all baselines in both environments, although for JacoRobot SIRL without VAE is better than with VAE.
In the GridRobot environment VAE pre-training helps SIRL.
However, while VAE performs comparably to other baselines in GridRobot, it severely underperforms in JacoRobot. This suggests that the reconstruction loss struggles to recover a helpful starting representation when the input space is higher dimensional and correlated. As a result, using the VAE pre-training to warmstart SIRL hinders performance when compared to starting from a blank slate. 
When comparing the preference-based baselines,
in GridRobot they all perform similarly apart from MultiPref with 50 humans. 
In JacoRobot we see a trend that more preference humans does not necessarily result in better performance. This confirms our observation from Figure  \ref{fig:FPE_simulation} that deciding on an appropriate number of human preferences to use for multi-task pretraining is challenging, a problem that SIRL bypasses. 

\smallskip
\noindent \textbf{Summary}. With enough representation data, SIRL 
learns representations more predictive of the true features (H1), leading to learning more generalizable rewards (H2). 
This does not necessarily mean that SIRL representations are perfectly aligned with causal features --- they are just \emph{better} aligned, so the learned rewards are also better.
When VAE pre-training recovers sensible starting representations it further reduces the amount of human data SIRL needs, otherwise it hurts performance. 
Lastly, surprisingly, Random is often better than pre-training with preference queries: preference-based methods may learn features that correlate with the training data but are not necessarily causal, and an incorrectly biased representation is worse for learning downstream rewards than starting from scratch.


\section{User Study}
\label{sec:study}

We now present a user study with novice users that provide similarity queries via an interface for the JacoRobot environment. 

\subsection{Experiment Design}

We ran a user study in the JacoRobot environment, modified for only two features: \textit{table} and \textit{laptop} (we removed the humanoid in the environment). We designed an interface where people can click and drag to change the view, and press buttons to replay trajectories and record their query answer (Figure \ref{fig:jacorobot_env}). We chose to display the Euclidean path of each trajectory in the query traces, as we found that to help users more easily compare trajectories to one another.

The study has two phases: collecting similarity queries and collecting preference queries. 
In the first phase, we introduce the user to the interface and describe the two features of interest.
Because similarity queries are preference-agnostic, we describe examples of possible preferences akin to the ones in Sec. \ref{sec:intro}, but we do not bias the participant towards any specific preference yet.
Each participant practices answering a set of pre-selected, unrecorded similarity queries, and then answers 100 recorded similarity queries. In the second phase, we describe a scenario in the environment that has a specific preference associated with it (e.g. ``There’s smoke in the kitchen, so the robot should stay high from the table'' or ``There is smoke in the kitchen and the robot’s mug is empty, so you want to stay far from the table and close to the laptop.'') and assign different preference scenarios to each participant. Each person practices unrecorded preference queries, then answers 100 preference queries.

\smallskip
\noindent \textbf{Participants.} We recruited 10 users (3 female, 6 male, 1 non-binary, aged 20-28) from the campus community to give queries. Most users had technical background, so we caution that our results will speak to SIRL’s usability with this population rather than more generally.

\smallskip
\noindent \textbf{Manipulated Variables.} Guided by the results in Figure  \ref{fig:TPA_simulation}, we compare our best performing method, SIRL without VAE, to Random, the best performing realistic baseline. For SIRL we collect 100 similarity queries from each participant and train a shared representation using all of their data.

\smallskip
\noindent \textbf{Dependent Measures.} We present the same two metrics from Sec. \ref{sec:experiments}, \textit{FPE} and \textit{TPA}. For \textit{TPA}, we collect 100 preference queries for each user's unique preference, we use 70\% for training individual reward networks which we evaluate on the remaining 30\% queries (Real). We compute \textit{TPA} with cross-validation on 50 splits. To demonstrate how well SIRL works for new people who don't contribute to learning the similarity embedding, we also train SIRL on the similarity queries of 9 of the users and compute \textit{TPA} on the held-out user's preference data (Held-out), for each user, respectively. Lastly, because real data tends to be noisy, we also compute \textit{TPA} with 70 simulated preference queries for 10 different rewards, which we also evaluate on a simulated test set (Simulated).

\smallskip
\noindent \textbf{Hypotheses.} Our hypotheses for the study are:

\textbf{H3}. Similarity queries (SIRL) recover more salient features than a random representation (lower \textit{FPE}), even with novice user data.

\textbf{H4}. The SIRL representation results in more generalizable reward learning (higher \textit{TPA}), even with novice similarity queries.

\begin{figure*}[t!]
\begin{subfigure}[b]{0.15\textwidth}
\centering
\includegraphics[width=\textwidth]{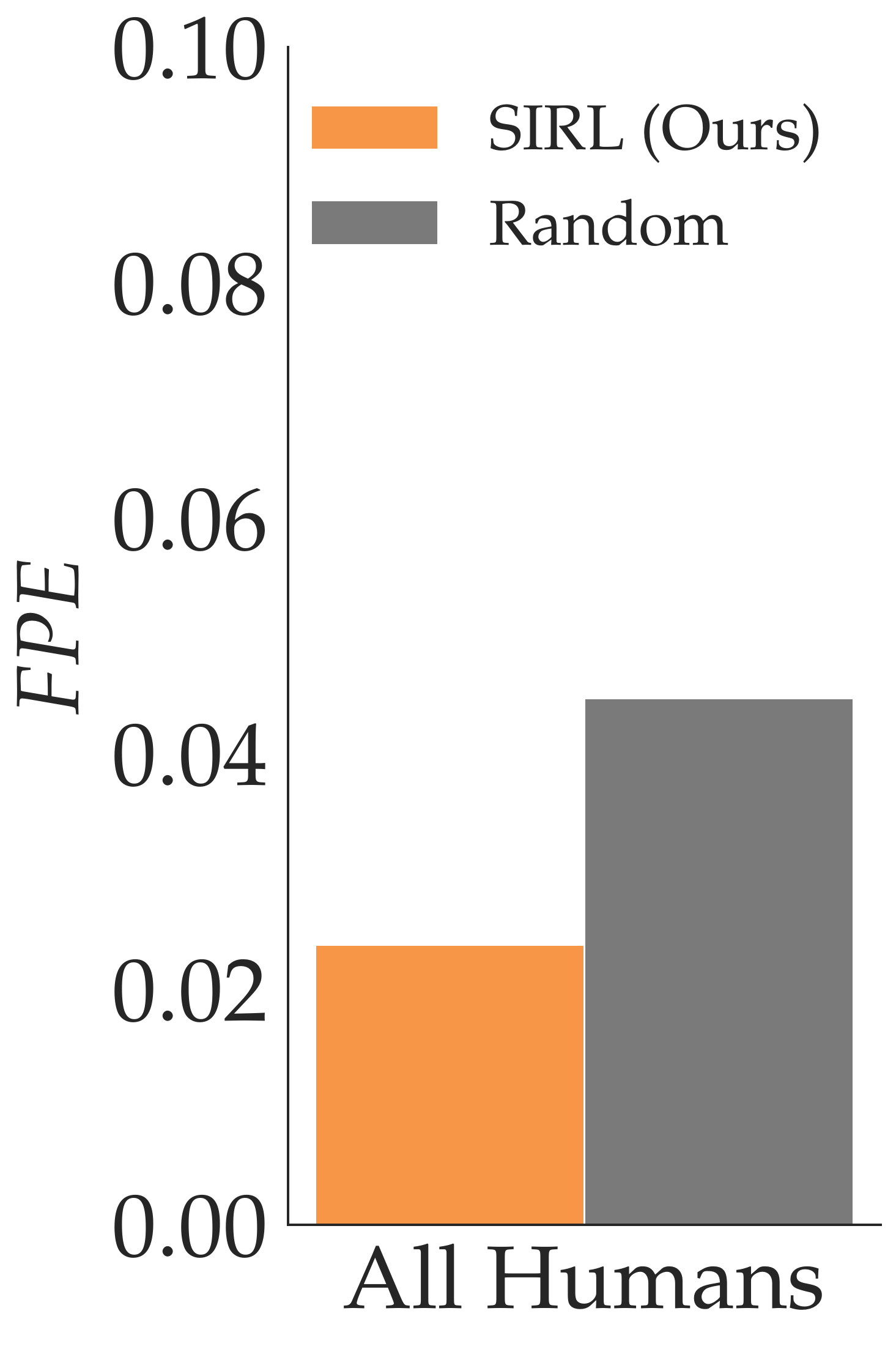}\\
 \label{fig:study_FPE}
\end{subfigure}
\hspace{1cm}
\begin{subfigure}[b]{0.635\textwidth}
\centering
\includegraphics[width=\textwidth]{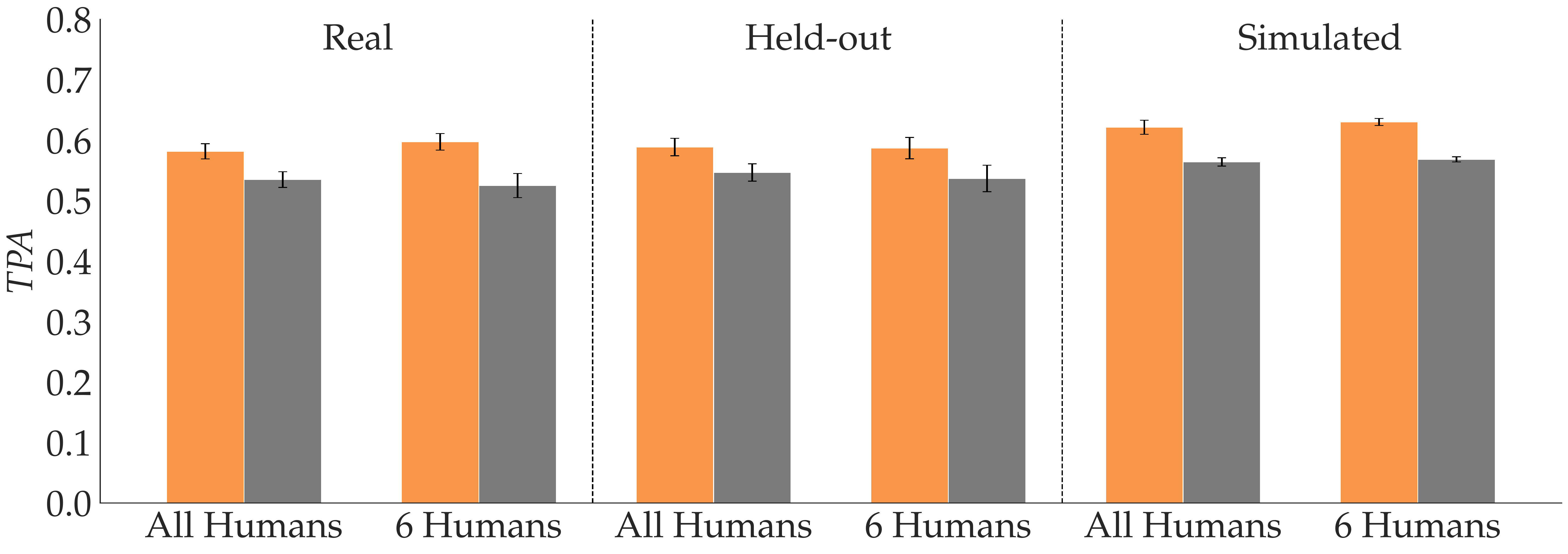}\\
 \label{fig:study_TPA}
\end{subfigure}
\caption{Study values for \textit{FPE}, and \textit{TPA} with real and simulated preferences. Even with novice similarity queries, SIRL recovers representations both more predictive of the true features and more useful for learning different user rewards than the baseline.}
\label{fig:study}
\vspace{-3mm}
\end{figure*}

\subsection{Analysis}

Figure  \ref{fig:study} summarizes the results. On the left, SIRL recovers a representation twice as predictive of the true features, supporting H3. A 2-sided t-test (p $<$ .0001) confirms this. This suggests SIRL can recover aspects of people's feature representation even with noisy similarity queries from novice users. On the right (Real), SIRL recovers more generalizable rewards on average than Random, providing evidence for H4. Furthermore, using the SIRL representation on a novel user (Held-out) also performs better than Random, and the result appears almost indistinguishable from Real. This  suggests that similarity queries can be effectively crowd-sourced and the resulting representation works well for novel user preferences. Lastly, training with simulated preference queries slightly improves performance for both methods, suggesting that noise in the human preference data can be substantial. Three ANOVAs with method as a factor find a significant main effect (F(1, 18) = 6.0175, p = .0246, F(1, 18) = 4.7547, p = .0427, and F(1, 18) = 16.1068, p $<$ .001, respectively). For each of the 3 cases, we also separated the 6 humans that were assigned preferences pertaining to both features (e.g. ``There is smoke in the kitchen and the robot’s mug is empty, so stay far from the table and close to the laptop.''). SIRL performance is slightly better when using preference data from this subset of users, hinting that perhaps the learned representation entangled the two features.

Overall, the quantitative results support H3 and H4, providing evidence that SIRL can recover more human-aligned representations. 
Subjectively, some users found the 2D interface deceiving at times, as they would judge trajectory similarity differently based on the viewpoint. This is a natural artifact of visualizing a 3D world in 2D, but future work should investigate better interfaces. Some users reported struggling to trade off the two features when comparing trajectories. This is in part due to the fact that we almost ``engineered'' their internal representation, so a more in-the-wild study could determine whether similarity queries are indeed preference-agnostic. Lastly, some queries were easier than others: users' time-to-answer varied across triplets suggesting that future work could use it as a confidence metric for how much to trust their answer. 

\section{Discussion and Limitations}
\label{sec:discussion}

In this work, our goal was to tackle the problem of learning good representations that capture (and disentangle) the features that matter, while excluding spurious features. If we had such a representation, then learning rewards that capture different preferences and tasks on top of it would lead to generalizable models that reliably incentivize the right behavior across different situations, rather than picking up on correlates and being unable to distinguish good from bad behaviors on new data. Our idea was to implicitly tap into this representation by asking people what they find similar versus not, because two behaviors will be similar if and only if their representations are similar. We introduced a novel human input type, trajectory similarity queries, and tested that it leads to better representations than those learned through self-supervision or via multi-task learning: it enables learning rewards from the same training data that better rank behaviors on test data. 

That said, we need to be explicit that this is not the be-all end-all solution to our goal above. The representations learned, as we see in simulation, are not perfectly aligned with causal features --- they are just \emph{better} aligned. The learned rewards are not perfect --- they are just better than alternatives. Similarity queries do not solve the problem fully, potentially because they suffer from the same issue preference queries do: when multiple important features change, it becomes harder to make a judgement call on what is more similar. The advantage that we see from similarity queries, though, is that rewards for particular tasks might ignore or down-weigh certain features that matter in other tasks, while similarity queries are task-agnostic and implicitly capture the distribution of tasks in the human's head. Rather than having to specify a task distribution for multi-task learning, with similarity queries we are (implicitly) asking the user to leverage their more general-purpose representation of the world. 
But thinking about how to overcome the challenge that multiple changing features make these queries harder to answer opens the door to exciting ideas for future work. For instance, what if we iteratively built the space, and based similarity queries on some current estimate of what are the important features; over time, as the representation becomes more aligned with the human's, the queries would get better at honing in on specific features. 

Another obvious limitation is that we did not do an in-the-wild study. In theory, similarity queries should be used when people already have a robot they are familiar with and, thus, have a distribution of tasks they care about in their everyday contexts, but in our study we needed to explain to users these contexts and what might be important. In doing so, we almost ``engineered'' their internal representation. As robots become more prevalent, a follow-up study where users are given much less structure and allowed to actually tap into \emph{their} unaltered representation might be possible.

In a sense, with SIRL we build a foundation model, and this sometimes requires hundreds of queries to learn a good representation. While we don’t think having this much data when pre-training is unreasonable, especially since it leads to significant desirable performance improvement over baselines, sample-complexity is crucial to address as we scale to more complex robotic tasks.
Because similarity queries are task agnostic, we can crowd-source the queries from multiple people (as we did in the user study) and rely on this economy at scale to alleviate user burden. Future work could also look at active querying methodologies to ask the person for more informative similarity queries and reduce data amounts.

A further avenue of work is extending this beyond reward learning, using SIRL representations directly for policy learning or learning exploration functions. We also emphasize that similarity queries are not a replacement for self-supervised learning; instead, we view them as complementary --- self-supervised learning might be able to leverage expert-designed heuristics to eliminate many of the spurious features, while SIRL might serve to fine-tune the representation. How to properly combined the two remains an open question.

Overall, similarity queries are a step towards recovering human-aligned representations. They improve upon the state of the art, and can benefit from further exploration in how to combine them with other inputs and self-supervision, and how to make them easier through better interfaces and query selection algorithms.


\bibliographystyle{ACM-Reference-Format}
\balance
\bibliography{0_HRI}

\clearpage
\nobalance
\appendix
\section{Appendix}

\subsection{Trajectory generation}
\label{app:trajgen}

In GridRobot the state space is discretized, so the trajectory space $\Xi$ can be enumerated; however, the JacoRobot state space is continuous,
so we need to construct $\Xi$ by sampling the infinite-dimensional trajectory space. We randomly sample 10,000 start-goal pairs and compute the shortest path in the robot's configuration space for each of them, $\xi^{SG}$. Each trajectory has a horizon length $H$ and consists of $n$-dimensional states. We then apply random torque deformations $u$ to each trajectory to obtain a deformed trajectory $\xi^{SG}_D$. In particular, we randomly select up to 3 states along the trajectory, and then deform each of the selected states with a different random torque $u$. To deform a trajectory in the direction of $u$ we follow: 
\begin{equation}
    \begin{aligned}
        \xi^{SG}_D = \xi^{SG} + \mu A^{-1} \tilde u
    \end{aligned}
    \enspace,
\end{equation}
where $A \in \mathbb{R}^{n(H+1) \times n(H+1)}$ defines a norm on the Hilbert space of trajectories and dictates the deformation shape \cite{deformation}, $\mu>0$ scales the magnitude of the deformation, and $\tilde u\in \mathbb{R}^{n(H+1)}$ is $u$ at indices $nt$ through $n(t+1)$ and 0 otherwise ($\tilde u$ is 0 outside of the chosen deformation state index). For each deformation, we randomly generated $\mu$ and the index of the state the deformation is applied to. For smooth deformations, we used a norm $A$ based on acceleration, but other norm choices are possible as well (see \citet{deformation} for more details). We took inspiration for this deformation strategy from \citet{bajcsy2017phri}.

\subsection{Training details}
\label{app:details}

We present architecture and optimization details that can assist in reproducing our training setup.

\subsubsection{Feature networks}

All embeddings have the same network size: for GridRobot we used MLPs with 2 hidden layers, 128 units each, mapping to 6 output neurons, while for JacoRobot we used 1024 units to handle the larger input space. For both environments, we used ReLU non-linearities after every linear layer.

We trained the VAE network with a standard variational reconstruction loss~\cite{kingma2014vae} also including a KL-divergence-based regularization term (to make the latent space regular). The regularization part of the loss had a weight of $\lambda=0.01$. For both environments, we optimized the loss function using Adam for 2000 epochs with an exponentially decaying learning rate of 0.01 (decay rate 0.99999) and a batch-size of 32. 

SinglePref and MultiPref with 10 and 50 humans are trained using the
standard preference loss in Eq. \eqref{eq:preference}. \citet{christiano2017preferences} ensured that the rewards predicted by the preference network remain small by normalizing them on the fly. We instead add an $l_2$ regulatization term on the predicted reward to the preference loss, with a weight of 10 for GridRobot and 1 for JacoRobot. All three methods optimize this final loss in the same way: for GridRobot, we use Adam for 5000 epochs with a learning rate of 0.01 and batch-size 32, while for JacoRobot we found a lower learning rate of 0.001 to result in more stable training.

Lastly, for SIRL we had the option to first pre-train with the above VAE loss. Training with the similarity objective in Eq. \eqref{eq:SIRLloss} happens disjointly, after pre-training. For both GridRobot and JacoRobot, we optimized this loss function using Adam for 3000 epochs with an exponentially decaying learning rate of 0.004 (decay rate 0.99999) and batch-size 64.

We note that our current architectures assume fixed-length trajectories but one could adopt an LSTM-based architecture for trajectories of varying length~\citep{sripathy2022emotive}.

\subsubsection{Preference networks}

For evaluating \textit{TPA}, we used preference networks on top of the embeddings for the respective methods we evaluate. 
For GridRobot we used MLPs with 2 hidden layers of 128 units, and for JacoRobot we used 1024 units for larger capacity. For both environments, we used ReLU non-linearities after every linear layer. We added the same $l_2$ regularization to the loss in Eq. \eqref{eq:preference} as before, with weight 10 for GridRobot and 1 for JacoRobot. For GridRobot, we optimized our final loss function using Adam for 500 epochs with a learning rate of 0.001 and a batch-size of 64. For JacoRobot, we increased the number of epochs to 1000.

\begin{figure*}[t]
\begin{subfigure}[b]{0.44\textwidth}
\centering
\includegraphics[width=\textwidth]{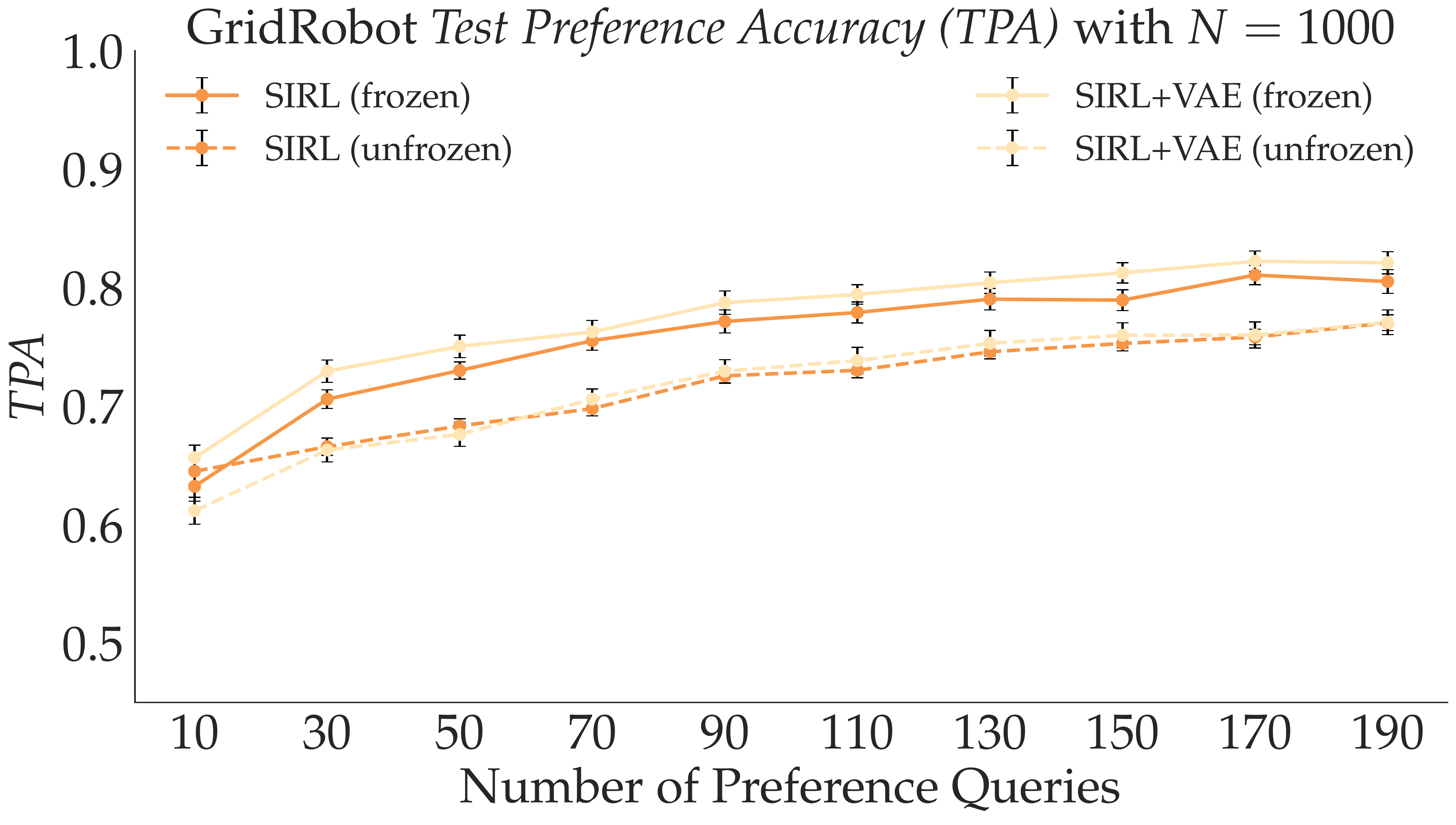}\\
\end{subfigure}
\begin{subfigure}[b]{0.44\textwidth}
\centering
\includegraphics[width=\textwidth]{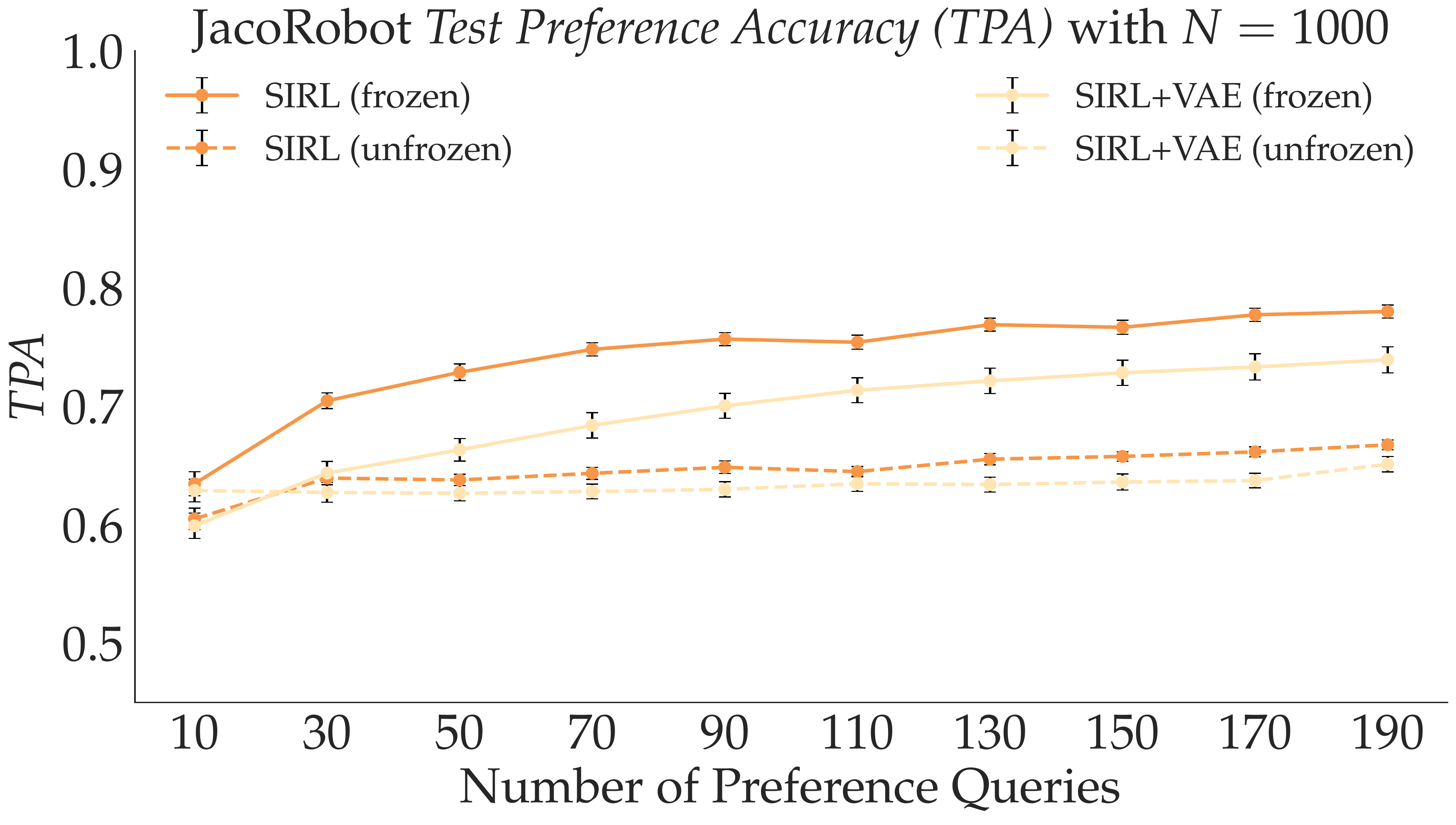}\\
\end{subfigure}
\begin{subfigure}[b]{0.44\textwidth}
\centering
\includegraphics[width=\textwidth]{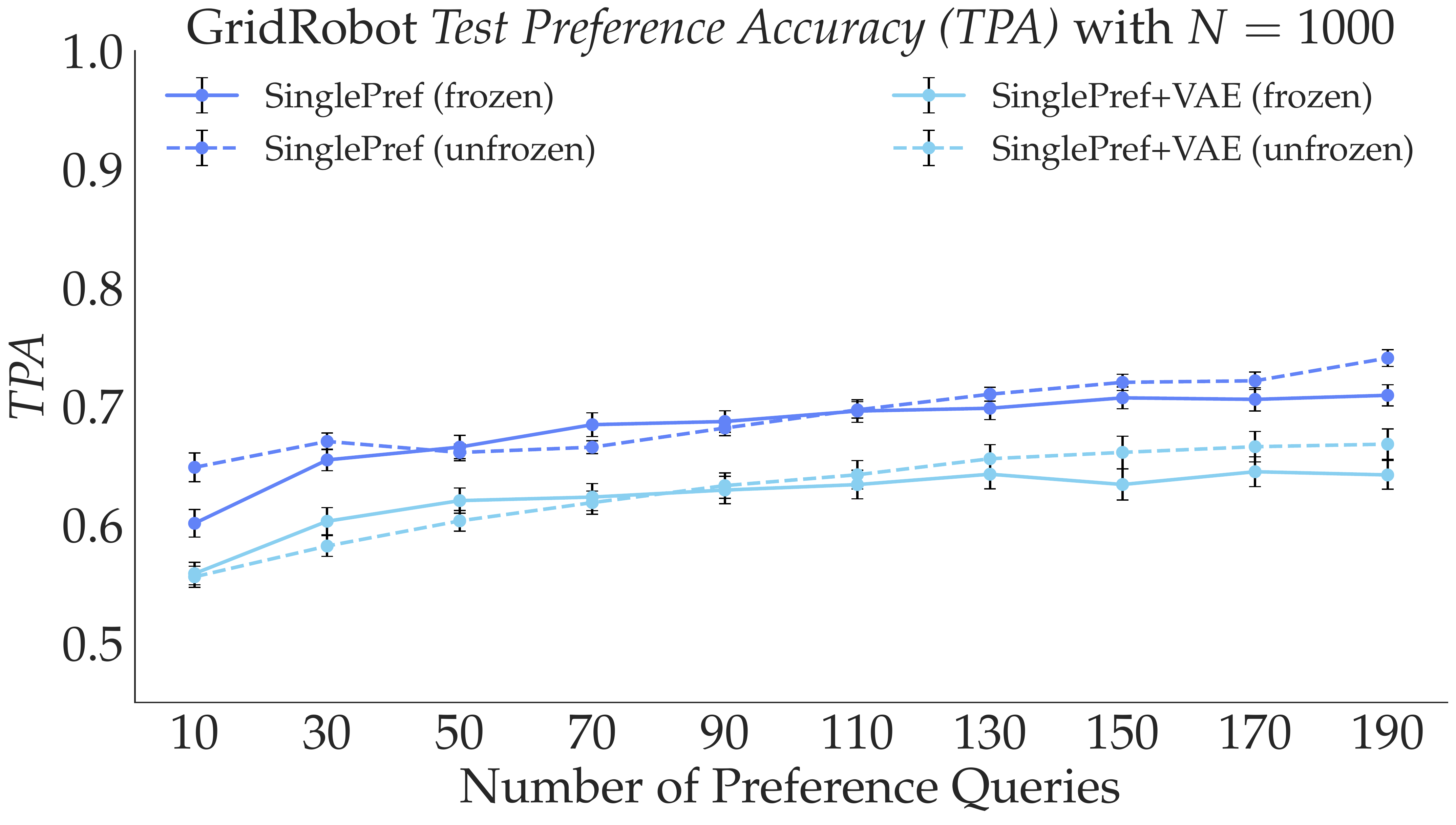}\\
\end{subfigure}
\begin{subfigure}[b]{0.44\textwidth}
\centering
\includegraphics[width=\textwidth]{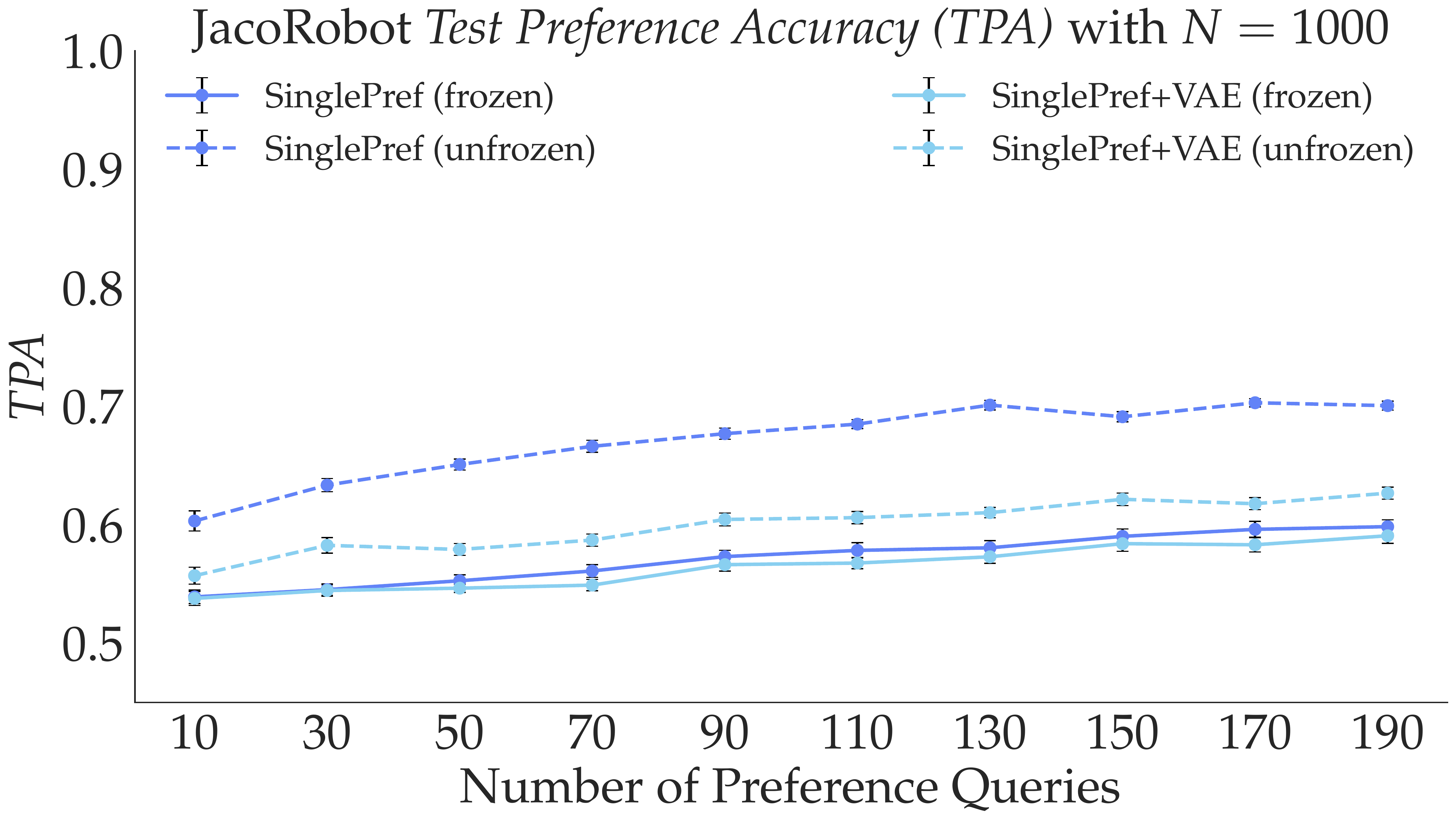}\\
\end{subfigure}
\begin{subfigure}[b]{0.44\textwidth}
\centering
\includegraphics[width=\textwidth]{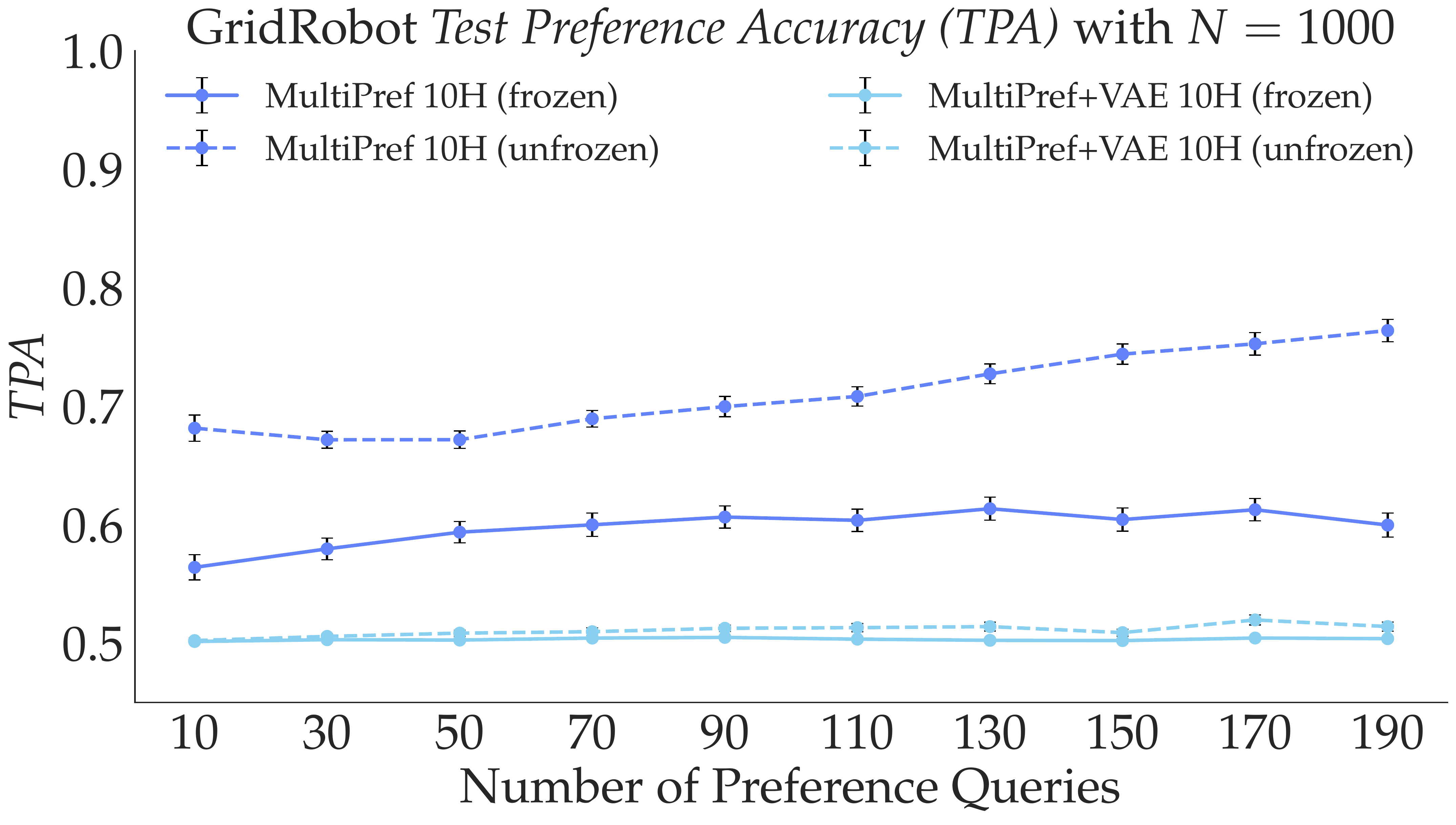}\\
\end{subfigure}
\begin{subfigure}[b]{0.44\textwidth}
\centering
\includegraphics[width=\textwidth]{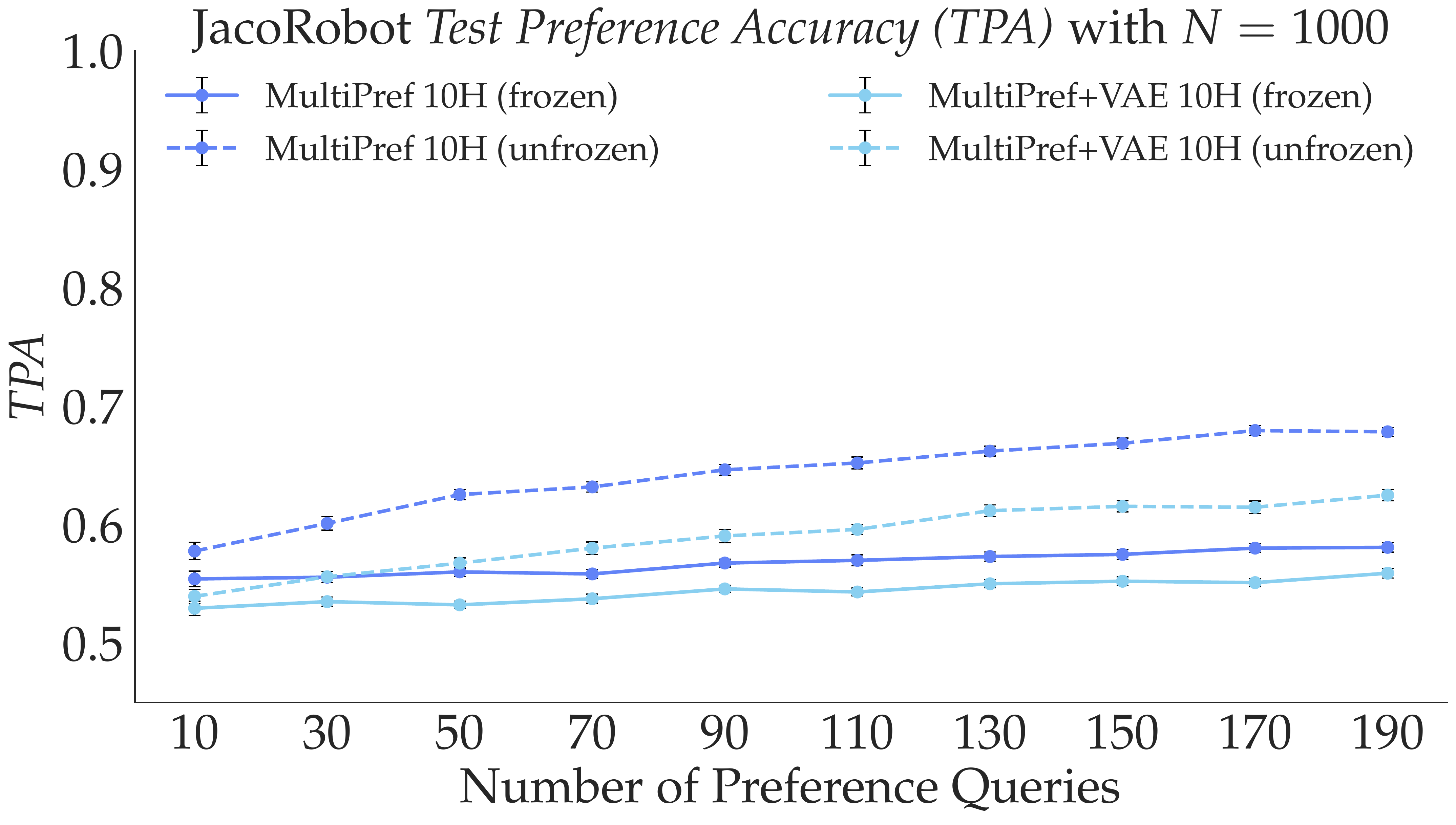}\\
\end{subfigure}
\begin{subfigure}[b]{0.44\textwidth}
\centering
\includegraphics[width=\textwidth]{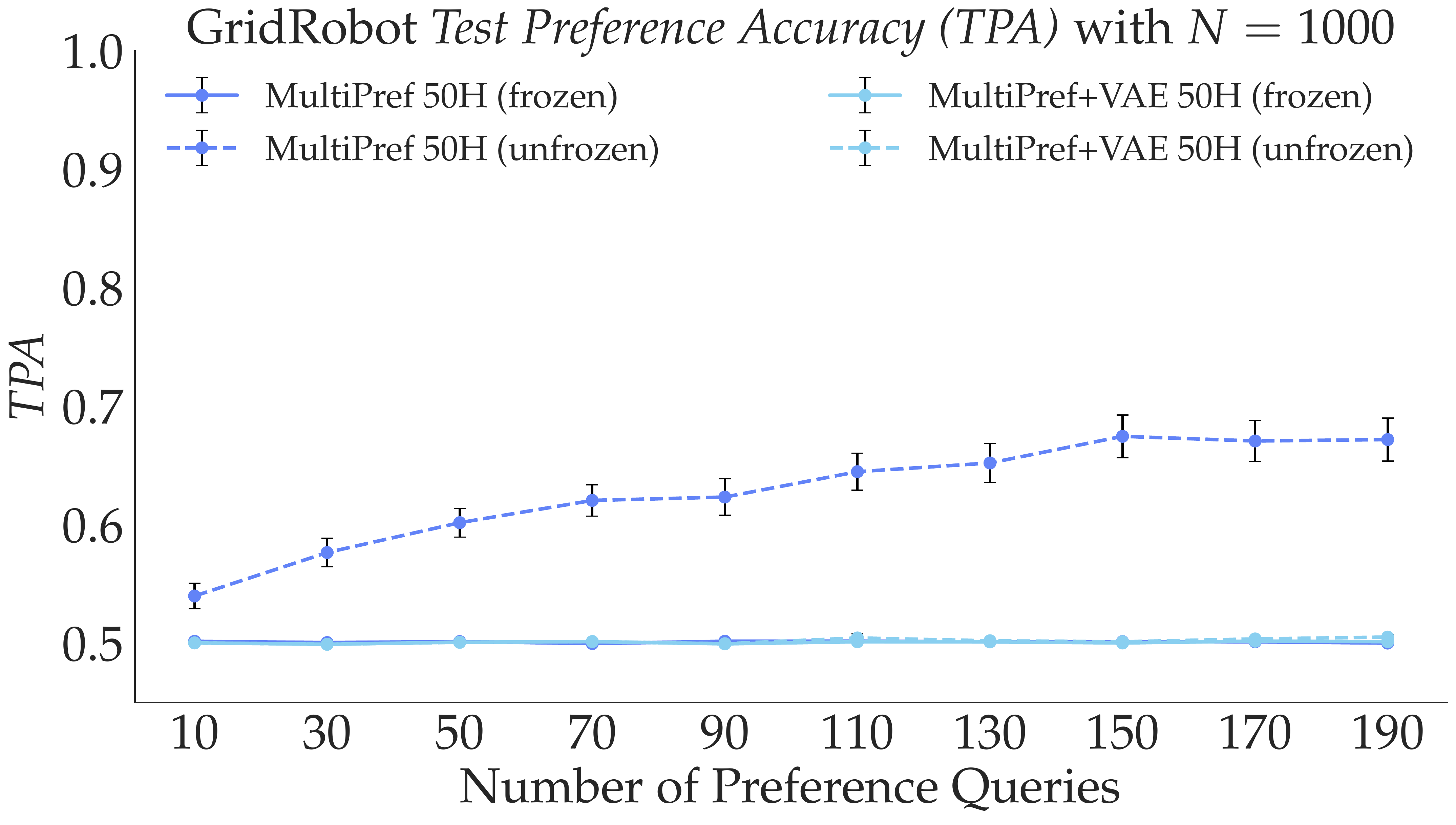}\\
\end{subfigure}
\begin{subfigure}[b]{0.44\textwidth}
\centering
\includegraphics[width=\textwidth]{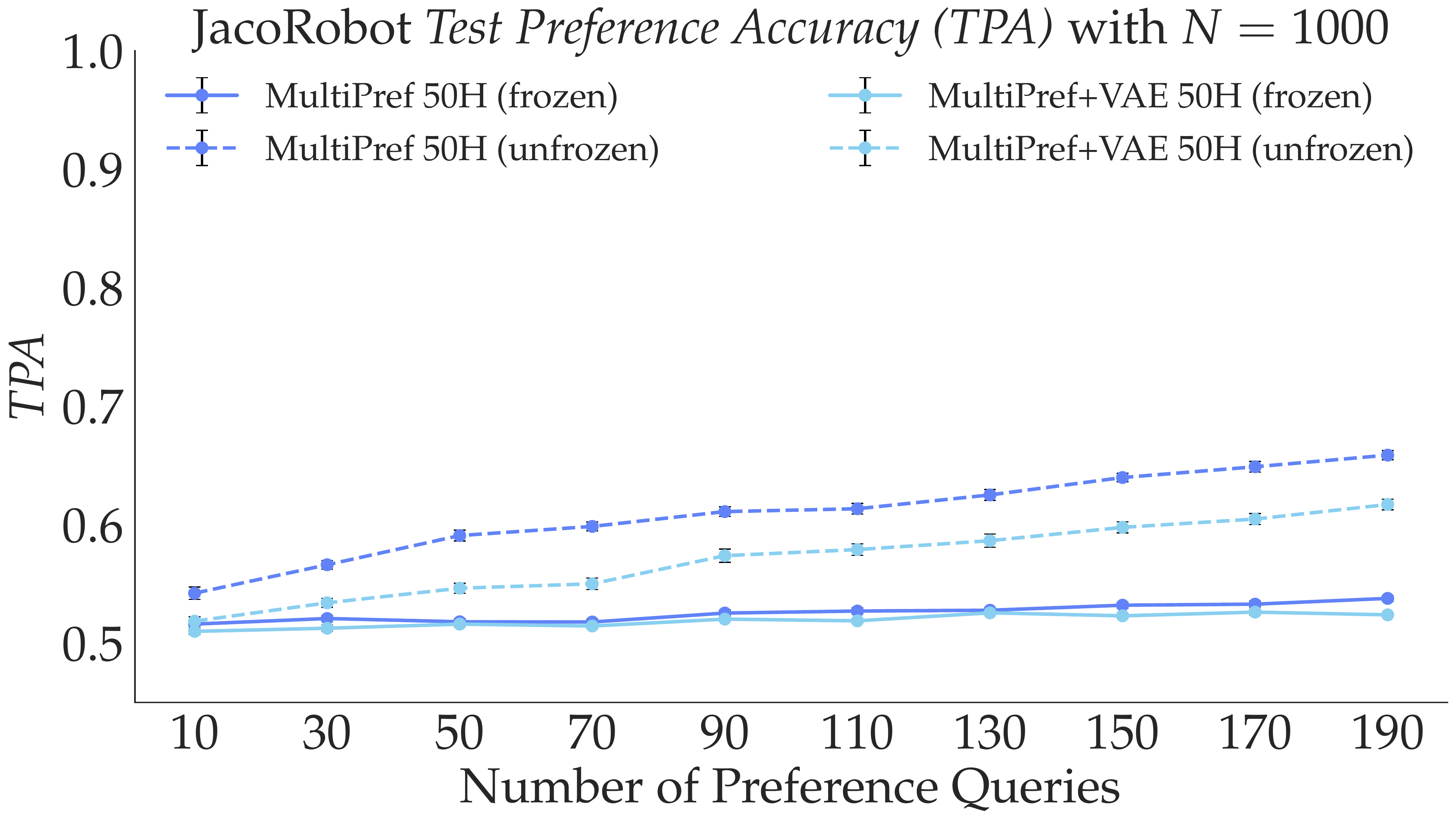}\\
\end{subfigure}
\begin{subfigure}[b]{0.44\textwidth}
\centering
\includegraphics[width=\textwidth]{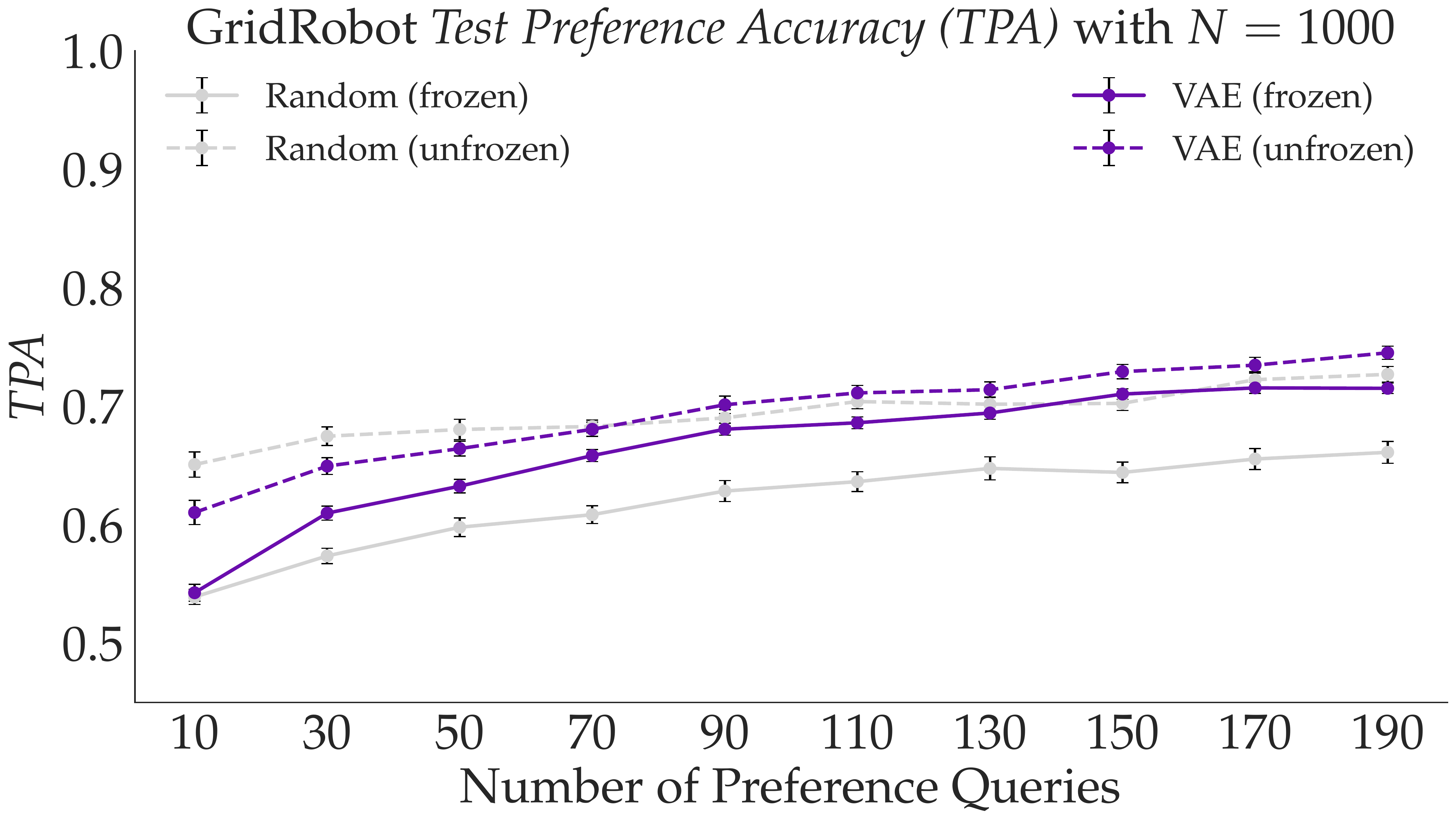}\\
\end{subfigure}
\begin{subfigure}[b]{0.44\textwidth}
\centering
\includegraphics[width=\textwidth]{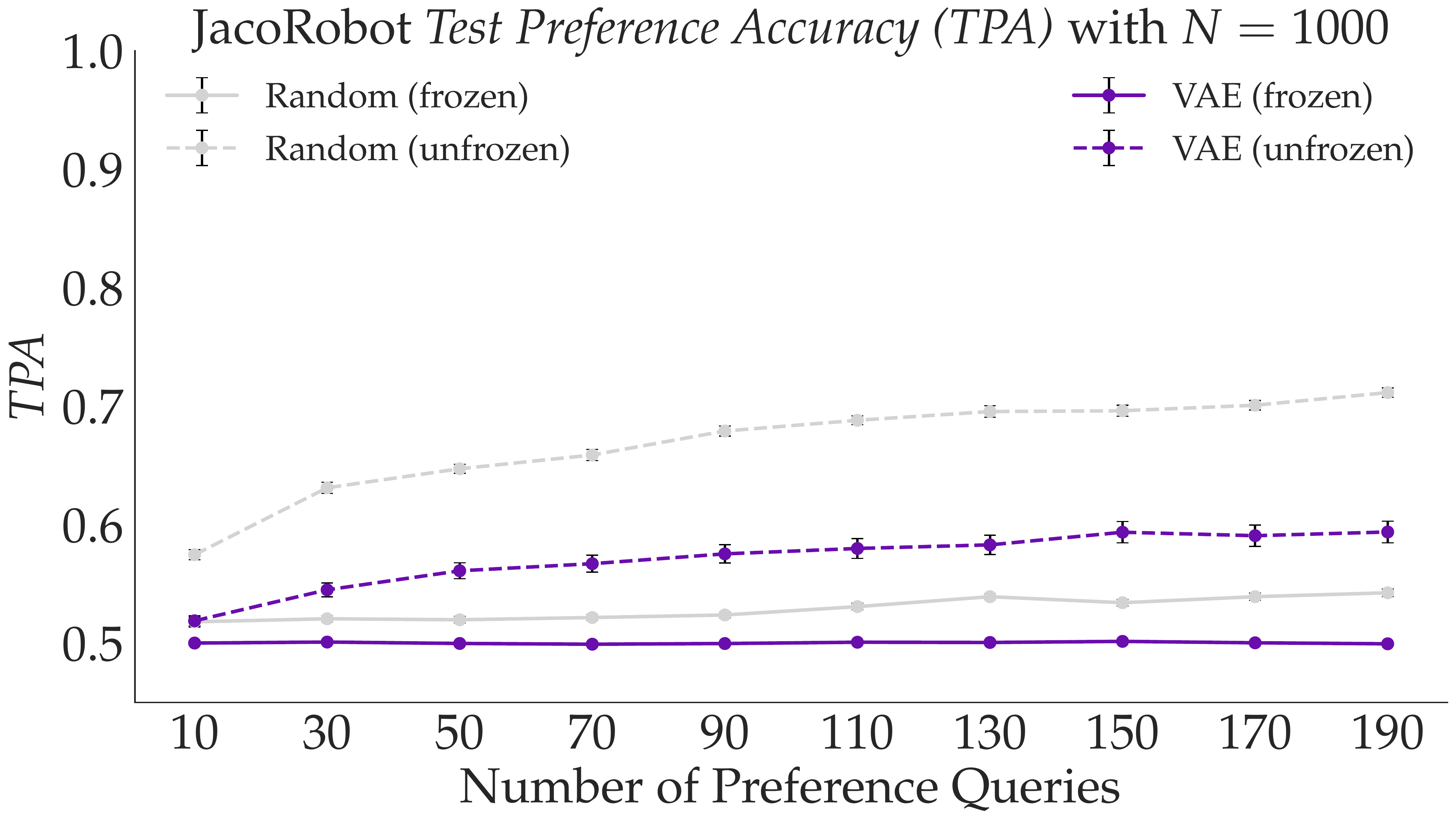}\\
\end{subfigure}
\caption{Ablation results for GridRobot (left) and JacoRobot (right). Overall, SIRL does better when the learned representation is frozen, while all other method do better when the representations is unfrozen. SinglePref and the MultiPref baselines perform better without VAE pre-training, while SIRL sometimes benefits from pre-training in simple environments like GridRobot.}
\label{fig:TPA_ablation}
\end{figure*}

\subsection{Ablations}
\label{app:ablations}

Figure \ref{fig:TPA_simulation} illustrates results with frozen SIRL, and unfrozen baselines without VAE pre-training, as these were the best configurations we found for each method. In this section, we show the complete ablation we performed to decide which methods benefit from frozen or unfrozen embeddings, or VAE pre-training. Figure \ref{fig:TPA_ablation} showcases the result of this ablation on both GridRobot and JacoRobot. Overall, we see that SIRL does better when the learned representation is frozen, while all the other methods do better when the representations is unfrozen. SinglePref and the MultiPref baselines perform better without VAE pre-training, while SIRL sometimes benefits from pre-training in simple environments like GridRobot.

\end{document}